\documentclass[letterpaper, preprint, paper,11pt]{AAS}
\usepackage{placeins}
\usepackage[utf8]{inputenc}
\usepackage{colortbl}
\usepackage{url}
\usepackage{amsmath} 
\usepackage{amssymb}  
\usepackage{graphicx}
\usepackage{hyperref}
\usepackage{mathtools}
\usepackage{amsthm}
\usepackage{xcolor}
\usepackage{bm}
\usepackage{graphicx}
\usepackage{subcaption}  
\usepackage{siunitx}
\usepackage{tabularx}
\usepackage{subcaption}

\newcommand{\R}{\mathbb{R}} %
\newcommand\largeblacktriangle{\scalebox{1.35}[1.25]{$\blacktriangle$}}
\newcommand{\acronym}{\textcolor{black}{AmorGS}}

\definecolor{DarkOrange}{RGB}{255, 140, 0}
\definecolor{Chartreuse}{RGB}{127, 255, 0}

\PaperNumber{23-352}

\title{
Amortized Global Search for Efficient Preliminary Trajectory Design with Deep Generative Models
}

\author{
Anjian Li\thanks{Ph.D. Candidate, Department of Electrical and Computer Engineering, Princeton University, NJ, USA.}, \ 
Amlan Sinha\thanks{Ph.D. Candidate, Department of Mechanical and Aerospace Engineering, Princeton University, NJ, USA.}, \ and
Ryne Beeson\thanks{Assistant Professor, Department of Mechanical and Aerospace Engineering, Princeton University, NJ, USA.}
}

\date{March 2023}

\begin{document}

\maketitle

\begin{abstract}

Preliminary trajectory design is a global search problem that seeks multiple qualitatively different solutions to a trajectory optimization problem.
Due to its high dimensionality and non-convexity, and the frequent adjustment of problem parameters, the global search becomes computationally demanding. 
In this paper, we exploit the clustering structure in the solutions and propose an amortized global search (\acronym{}) framework.
We use deep generative models to predict trajectory solutions that share similar structures with previously solved problems, which accelerates the global search for unseen parameter values.
Our method is evaluated using De Jong's 5th function and a low-thrust circular restricted three-body problem.

\end{abstract}

\section{Introduction}


In the preliminary mission design phase, the low-thrust spacecraft trajectory design is a global search problem \cite{yam2011low, englander2012automated}. 
Mission designers attempt to find a rich collection of qualitatively different solutions to a trajectory optimization problem, in order to make tradeoffs between various mission objectives and requirements.

Solving this low-thrust trajectory optimization problem is a challenging task due to its high dimensionality and non-convexity, which result in many local optima.
To conduct a global search for locally optimal solutions, it is essential to thoroughly explore the variable space and solve this high-dimensional optimization problem from different initializations.
However, without \textit{a priori} knowledge of the specific solution structures, exploring the entire space is extremely time-consuming.
On the other hand, solving a high-dimensional optimization problem within a limited time is inherently challenging. 
Consequently, the complexity of the dynamical model and the degrees of freedom of the optimization problems are often restricted in the preliminary phase. 

Reducing the dynamical model complexity alleviates the non-convexity of the optimization problems, and reducing the degrees of freedom (e.g., in control transcriptions) lowers the dimensionality of the optimization problem. 
These reductions allow for a quicker global search but at the risk that solutions may become infeasible when re-evaluated at flight fidelity. 
Therefore, there is a balance to be struck between the fidelity of the models used in the preliminary design phase and the extensiveness of the global search. 

Moreover, the global search problem usually encompasses many different parameters that shape the trajectory design.
Common examples include spacecraft system parameters, departure and arrival time, boundary conditions, and mission objectives, to name a few. 
In response to the evolving design requirements in the preliminary design phase, the values of these parameters are often subject to frequent adjustments.
As a result, we need to repetitively conduct the expensive global search over similar trajectory optimization problems with varying parameter values.

In this paper, we uncover the structure of the local optimal solutions to classic low-thrust trajectory optimization problems.
These solutions appear to be grouped into clusters, supporting long-standing hypotheses from Yam et al. \cite{yam2011low} and Englander et al. \cite{englander2012automated}.
We also discover that for similar trajectory optimization problems, which differ only in their parameter values, e.g., maximum allowable thrust for the spacecraft, locally optimal solutions display similar clustering behaviors.
Drawing from these observations, we propose a data-driven method to predict the solution structure of an unseen low-thrust trajectory optimization problem based on pre-solved similar problem instances.
Then the global search for this new problem can be warm-started from structured predictions instead of relying solely on the random exploration of the variable space.
This approach largely accelerates the global search process, focusing on the area that is more likely to have local optima.
Looking ahead, we may be able to allow for higher fidelity solutions in earlier design phases while still producing a sufficiently thorough global search. 

\subsection{Problem formulation}

The original low-thrust trajectory optimization problem is a continuous time optimal control problem.
In this study, we use the direct approach to convert the optimal control problem to a discretized nonlinear program (NLP). 
Here, a control transcription such as forward-backward shooting, as mentioned in \cite{beeson_dylan_2022}, is used to convert the control variable from a path into a finite-dimensional space of $\R^n$.
It's important to note that the trajectory optimization problem also depends on certain problem parameters, such as spacecraft system parameters.
Hence we formulate a family of parameterized optimization problems $\mathcal{P}_{\bm{\alpha}}$ as follows.
\begin{align} \label{eq: parameterized optimization}
    \mathcal{P}_{\bm{\alpha}} \coloneqq 
    \begin{cases}
    \underset{\bm{x}}{\min} \quad &J(\bm{x};\bm{\alpha})  \\
    s.t., \quad &c_{i}(\bm{x};\bm{\alpha}) \leq 0, \ \forall i \in \mathcal{I} \\
    \quad &c_{i}(\bm{x};\bm{\alpha}) = 0, \ \forall i \in \mathcal{E}
    \end{cases}  
\end{align} 
where $\bm{x} \in \mathcal{U} \subseteq \R^n$ is the decision variable, $\bm{\alpha} \in \R^k$ represents the problem parameter,
$J \in C^2(\mathcal{U}, \R^k; \R)$ is the objective function that maps from the variable $\bm{x}$ and parameter $\bm{\alpha}$ to the objective value, and $c_i \in C^2(\mathcal{U},\R^k; \R)$ are a set of inequality when $i \in \mathcal{I}$ and equality constraints when $i \in \mathcal{E}$.
$J$ and $c_i$ are assumed to be nonlinear functions.
In the context of the trajectory design problem, $\bm{x}$ can be the control input and time-of-flight; $\bm{\alpha}$ are the system parameters (e.g., maximum allowable thrust, limits on the control inputs, departure and arrival time, boundary conditions, etc); $J$ is the objective function to be minimized (e.g., fuel expenditure, etc.), and $c_i$'s are the system constraints (e.g., the equations of motion the spacecraft needs to satisfy, etc.).

We assume that each optimization problem in $\mathcal{P}_{\bm{\alpha}}$ shares the same problem structure (i.e., the same function $J$ and $c_i$), only with different problem parameter values of $\bm{\alpha}$.
The primary goal of our work is, given a new set of parameter value $\bm{\alpha' }$, to conduct an efficient global search for a rich collection of local optimal solutions $\{\bm{x}^{*}_i(\bm{\alpha' })\}_{i=1}^{N}$ to the problem $\mathcal{P}_{\bm{\alpha' }}$.

\subsection{Background and Related Literature}

For a specific trajectory optimization problem $\mathcal{P}_{ \bm{\alpha}}$ with parameter $ \bm{\alpha}$, 
numerous studies have focused on global search methods that identify high-quality and qualitatively different solutions.
For example, a grid-based search is a classical approach for spacecraft preliminary trajectory design.
However, this technique is more suitable for impulsive trajectory since the search space is much smaller.
Due to the curse of dimensionality, low-thrust trajectory design often needs a more intelligent global search algorithm.
Evolutionary algorithms, including Differential Evolution (DE) \cite{price2006differential}, Genetic algorithm (GA) \cite{mitchell1998introduction}, Particle swarm optimization (PSO) \cite{kennedy1995particle}, etc., have been widely used in global optimization problems in spacecraft trajectory design \cite{izzo2007search, vinko2007benchmarking, izzo2010global, vasile2010analysis}.
These algorithms iteratively generate new solutions by introducing randomness to previously obtained solutions and downselecting the solutions based on specific quality metrics.
In addition, researchers also combine stochastic search algorithms with local gradient-based optimizers to attempt to find the globally optimal solution.
The multistart method samples the search space with a fixed distribution and feeds the samples into a local optimizer as starting points for local search \cite{vasile2010analysis}.
Inspired by energy minimization principles in computational chemistry, Monotonic Basin Hopping (MBH) \cite{wales1997global, leary2000global} adds random perturbations during the local search to uncover multiple local optima solutions that are close to each other. 
MBH rapidly became popular in the sphere of spacecraft trajectory design \cite{yam2011low, addis2011global, englander2017automated} and has been established as the state-of-the-art algorithm in terms of efficiency and solution quality through various benchmarks \cite{vasile2008testing, izzo2010global, vasile2010analysis}.

Solving $\mathcal{P}_{\bm{\alpha}}$ entails providing an initial guess $\bm{x}^0$ for the solution to a local optimizer, which is then iterated upon to arrive at a locally optimal solution.
When a new problem $\mathcal{P}_{\bm{\alpha}'}$ is given with \textit{a priori} unknown parameter $\bm{\alpha}'$, to explore the solution space, MBH usually relies on a fixed distribution from which the initializations $\bm{x}^0$ are sampled, such as the uniform distribution \cite{englander2017automated}.
To improve the robustness and efficiency of the MBH, Englander et al. \cite{englander2014global, englander2014tuning} choose Cauchy or Pareto distributions  for the long-tailed distributions.
However, sampling the initial guesses $\bm{x}^0$ from only fixed distributions cannot consider the specific structure in high-dimensional solutions, and often requires tedious hand-tuning of the distribution parameters.
When repetitively solving problems in Eq. \eqref{eq: parameterized optimization}, many of which may be very similar, MBH always starts from scratch for the problem with new parameters, neglecting the data generated from the previous solving process. 

Machine learning techniques have also been used in the global search for spacecraft trajectory optimization problems.
Cassioli et al. \cite{cassioli2012machine} use the Support Vector Machine (SVM)  to learn the mapping between the initialization and the corresponding final outcome from the local optimization, identifying favorable initial guesses $\bm{x}^0$.
But in this work, they only consider relatively simple optimization problems with box constraints or linear constraints, where finding locally optimal solutions is much easier.
Also, this SVM method tends to work better with low-dimensional problems,  otherwise, it would be difficult to densely sample the high-dimensional space and let SVM accept or reject the initial guesses.
Izzo et al. \cite{izzo2016designing} propose a tree search algorithm that has been used to explore the search space of trajectory solutions efficiently.
However, these methods have been only tested on the same problem as the training data, leaving their generalizability to new problems unverified.
Moreover, traditional machine learning methods like SVM lack expressiveness for complex data structures and cannot benefit from more training data.
On the contrary, Deep Neural Networks (DNN) is a powerful function approximation, which could potentially learn more effective representations from complex data and offer better scalability to a large number of high-dimensional inputs.

Recently, there has been a growing interest in amortized optimization, which leverages the observations that similar problem instances often share similar solution structures \cite{amos2022tutorial}. 
The key idea is to predict solutions to new optimization problems based on similar problems that are previously solved.
This data-driven method has been successfully applied to accelerate various optimization problems in the area of vehicle dynamics and control \cite{chen2022large} and portfolio optimization \cite{sambharya2022end}.
They both leverage the power of Deep Neural Networks to predict solutions, such as Multilayer Perceptrons (MLP) and Convolutional Neural Networks (CNN).
However, most existing amortized optimization methods focus on predicting a specific solution to the problem, such as the globally optimal solution.
This is typically achieved by learning a mapping from the problem parameter to an initialization \cite{chen2022large} or a solving strategy \cite{cauligi2021coco} that leads to that particular solution.
On the contrary, the trajectory design is a global search problem.
Our goal is to uncover a collection of locally optimal and qualitatively different solutions $\{\bm{x}^{*}_i(\bm{\alpha' })\}_{i=1}^{N}$ rather than merely pursuing the one with the globally optimal objective value.
Hence, we require a sampler to generate local optimal solutions from the distributions.

Generative models are designed to model the probability distribution of the input data and provide easy sampling for this distribution in a high-dimensional space.
Variational Autoencoders (VAE) \cite{kingma2013auto} are a widely used generative model with an encoder-decoder structure parameterized by deep neural networks.
Compared to another popular choice, Generative Adversarial Networks (GAN) \cite{goodfellow2014generative}, VAEs are renowned for their stable training process.
The VAE first encodes the input data to a latent variable with a simple distribution like Gaussian distribution, then decodes the samples from the latent space to generate samples in the original space that closely resemble the input distribution.
Especially, when the input data exhibits clustering behaviors, Jiang et al. \cite{jiang2016variational} adopt a Gaussian Mixture Model (GMM) as the prior distribution of the latent variable in the VAE to generate more structural outputs.
This method has proven to have great performance in image generation by labels \cite{jiang2016variational, dilokthanakul2016deep}, single-cell sequence clustering \cite{xiong2019scale}, etc.
Diffusion models \cite{sohl2015deep} have recently emerged as a promising class of generative models, which have demonstrated impressive performance in image synthesis \cite{ho2020denoising, rombach2022high} and trajectory generation \cite{ajay2022conditional, botteghi2023trajectory}. 
However, the theoretical understanding and practical application of diffusion models are still rapidly developing areas of research. Although this approach lies beyond the scope of the current paper, we plan to delve deeper into it in future work.

The Conditional VAE (CVAE) developed by Sohn at al. \cite{Sohn2015learning} introduces a conditional variable as an additional observation in the input to the VAE model.
This structure allows the CVAE to manipulate its sample generation by using different conditional observations, enabling the generalization of the sampling process to previously unseen observations.
For example, in the area of molecular biology, CVAE is used to generate drug-like molecules that satisfy certain target properties \cite{lim2018molecular}.
In autonomous driving scenarios, CVAE predicts the future trajectories of surrounding vehicles based on their trajectory histories \cite{salzmann2020trajectron++}.
However, to the best of our knowledge, no connection has been made between conditional generative models and amortized optimization, where the distribution of locally optimal solutions could be inferred and sampled based on the optimization problem parameters.

\subsection{Our Contributions} 

In this paper, we investigate the structure of locally optimal solutions $\bm{x}^{*}(\bm{\alpha})$ to low-thrust trajectory optimization problems.
We look into a minimum fuel low-thrust transfer in Circular Restricted Three-Body Problem (CR3BP) as an example $\mathcal{P}_{\bm{\alpha}}$, where parameter $\bm{\alpha}$ is the maximum allowable thrust for the spacecraft.
This $\bm{\alpha}$  can be replaced by other system parameters such as boundary conditions, departure and arrival time, etc., without the need to make additional changes.
When solving this problem $\mathcal{P}_{\bm{\alpha}}$ from uniformly sampled initializations, we find out that the locally optimal solutions $\bm{x}^{*}(\bm{\alpha})$ often fall into several clusters.
Each cluster contains qualitatively different trajectory solutions.
For instance, trajectories within some clusters may have more revolutions compared to other clusters.
Furthermore, when we conduct a global search for similar problem instances $\mathcal{P}_{\bm{\alpha}}$ with varying parameters $\bm{\alpha}_1, \bm{\alpha}_2, ...$, their corresponding local optimal solutions display similar clustering patterns.
This insight enables us to predict the solution structure for new problems  $\mathcal{P}_{\bm{\alpha}'}$ with unknown parameter values $\bm{\alpha}'$.
\FloatBarrier
\begin{figure}[!htbp]
    \centering
    \includegraphics[width=0.75\linewidth]{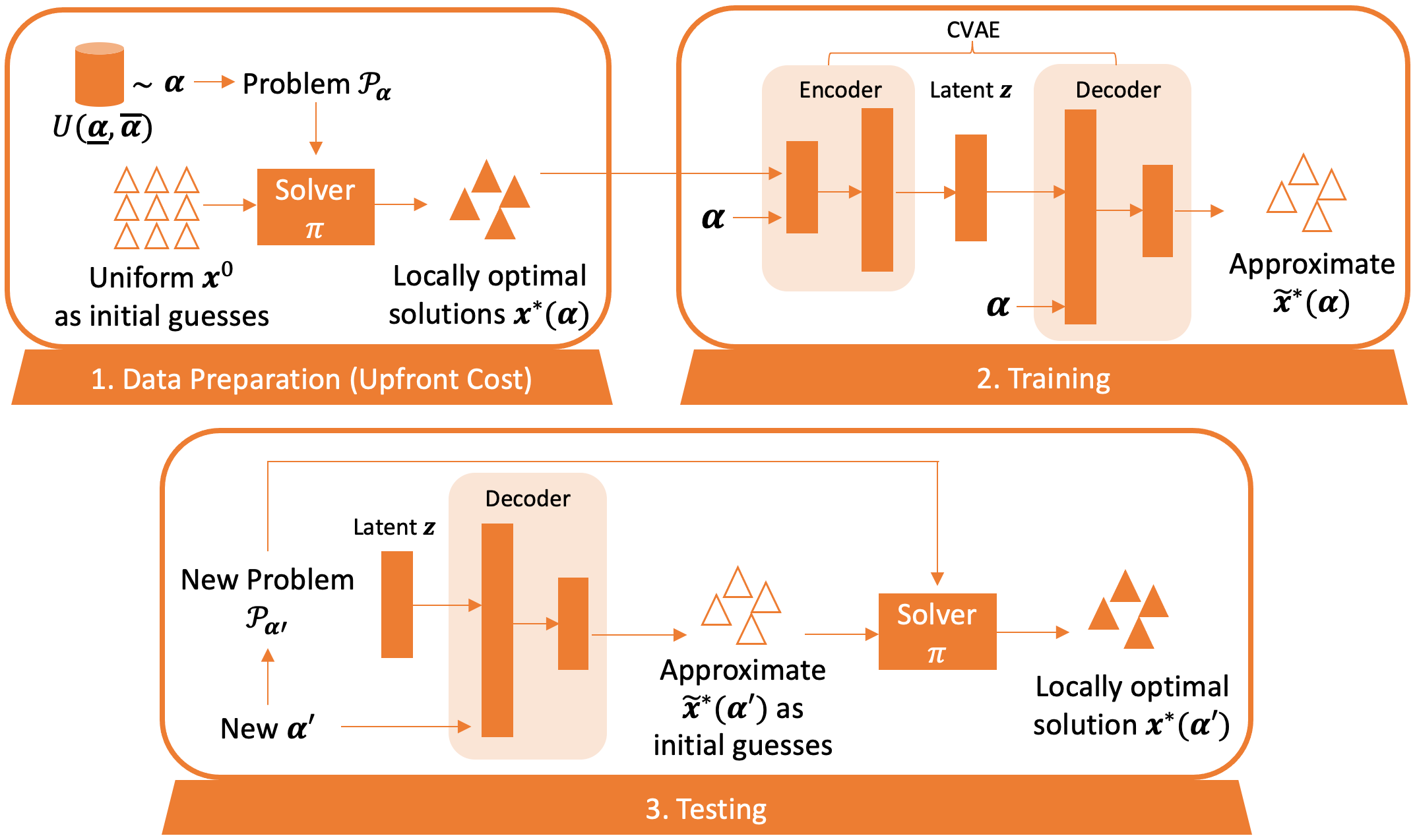}
    \caption{The workflow of \acronym{} framework to accelerate the global search over parameterized optimization problem $\mathcal{P}_{\bm{\alpha}}$.
    $\triangle$: initial guesses, $\largeblacktriangle$: final solutions.
    We first collect locally optimal solutions $\bm{x}^*(\bm{\alpha})$ to various problem $\mathcal{P}_{\bm{\alpha}}$ using a numerical local optimization solver $\pi$ with uniformly sampled initializations $\bm{x}^0$.
    Then we train a CVAE to generate the approximate solution $\bm{\tilde x}^*(\bm{\alpha})$ conditioned on $\bm{\alpha}$ with a similar structure as $\bm{x}^*(\bm{\alpha})$.
    When a new problem $\mathcal{P}_{ \bm{\alpha'}}$ is presented, we use the decoder to predict the structural approximated solution $\bm{\tilde x}^*(\bm{\alpha'})$, which then warm-start the global search to obtain the final collection of solutions $\bm{x}^*( \bm{\alpha'})$.
    }
    \label{Fig: work flow}
\end{figure}
\FloatBarrier
Given the aforementioned findings, we propose a novel framework \acronym{}, an amortized global search framework capable of predicting a collection of structured solution $\{\bm{x}^{*}_i(\bm{\alpha})\}_{i=1}^{N}$ to a trajectory optimization problem $\mathcal{P}_{\bm{\alpha}}$ based on the parameter value $\bm{\alpha}$.
In a data-driven manner, \acronym{}  utilizes the CVAE to predict the solution structure for a new problem leveraging similar problem instances that are previously solved.
To model the clustering behavior of the local optimal solutions, we select the GMM as the prior distribution of the latent variable.
The conditional input of the CVAE is the problem parameter that enables the generalization to unseen values.
The key insight is that trained on many solution data $\bm{x}^{*}(\bm{\alpha})$ with various parameter values $\bm{\alpha}$, the CVAE is capable of learning how the distribution of locally optimal solutions will change with respect to the parameter $\bm{\alpha}$, for a family of problems $\mathcal{P}_{\bm{\alpha}}$.
To the best of our knowledge, this is the first work that connects the idea of generative models and amortized optimization to accelerate global search on non-convex optimization problems. 

Our \acronym{} framework demonstrates a significant improvement in the computational efficiency of the global search over trajectory optimization problems $\mathcal{P}_{\bm{\alpha}}$.
Instead of a random exploration from scratch, the global search can be warm-started from the predictions generated by the CVAE. 
If the predictions are accurate, they will be close to the true optimal solutions from which it takes very few steps to converge. 
We validate the solution prediction performance of our proposed CVAE on a global optimization test function, De Jong's 5th function.
We also demonstrate our framework for low-thrust trajectory design in the CR3BP.
We show a large improvement in the efficiency of global search over this problem through warm-starting using the \acronym{} framework.
The efficiency can be further improved by combining our method with a more refined local search, such as the state-of-the-art MBH method.

\section{Methodology}


In this paper, we consider a class of problem  $\mathcal{P}_{\bm{\alpha}}$ where the locally optimal solutions fall into different clusters,  and each cluster contains a number of neighboring local optima. 
We will demonstrate in the experiment section that local optimal solutions to a low-thrust transfer in the CR3BP will exhibit such a clustering pattern. 
We also assume that the clusters of locally optimal solutions change smoothly with the problem parameter $\bm{\alpha}$.
This follows directly from the definition of $\mathcal{P}_{\bm{\alpha}}$ in Eq. \eqref{eq: parameterized optimization}, where both $f$ and $g_i$ are assumed to be $C^2$-differentiable with respect to $\bm{\alpha}$.
As a result, we assume that the clusters of locally optimal solutions do not change drastically with respect to small perturbations of $\bm{\alpha}$, allowing us to predict the locations of the solution clusters for new $\bm{\alpha}$.

\subsection{\acronym{} framework}

The \acronym{} framework aims to predict the locally optimal solutions based on similar problem instances that are previously solved.
In this paper, similar problem instances are those that belong to the same problem family $\mathcal{P}_{\bm{\alpha}}$, as outlined in Eq. \eqref{eq: parameterized optimization}, but vary in terms of their parameters $\bm{\alpha}$.
With these predictions as initial guesses, the global search can be warm-started in new problem instances.

The main workflow of \acronym{} is summarized in Figure \ref{Fig: work flow}.
Here we use the symbol $\triangle$ to represent initial guesses and $\largeblacktriangle$ to denote the corresponding final solutions obtained by a numerical solver $\pi$.
In the data preparation stage, we collect a set of locally optimal solutions $\bm{x}^*(\bm{\alpha})$ for various $\bm{\alpha}$ values drawn from a uniform distribution $\bm{\alpha} \sim U(\underline{\bm{\alpha}}, \overline{\bm{\alpha}})$.
We attain these solutions by solving different instances of $\mathcal{P}_{ \bm{\alpha}}$ in Eq. \eqref{eq: parameterized optimization} with a numerical local optimization solver $\pi$ and uniformly sampled initializations $\bm{x}^0$.
As many of these problems have been previously solved during the preliminary mission design phase, this data collection process will not add significant overhead.
During the training time, we train a CVAE model to take the locally optimal solution $\bm{x}^*(\bm{\alpha})$ and the problem parameter $\bm{\alpha}$ as inputs and generate samples $\bm{\tilde x}^*(\bm{\alpha})$ with a distribution approximating the input.
The CVAE is not explicitly modeling the distribution of the local optimal solutions $\bm{x}^*(\bm{\alpha})$, but rather provides an efficient sampling scheme to generate data that preserves a similar structure.
When a new problem $\mathcal{P}_{ \bm{\alpha'}}$ arises during testing, we can input the new parameter value $\bm{\alpha}'$ into the decoder of the CVAE and predict the approximate solutions $\bm{\tilde x}^*(\bm{\alpha'})$.
Finally, these predictions are used as the initialization $\bm{x}^0( \bm{\alpha'})$ for the solver $\pi$ to solve the optimization problem $\mathcal{P}_{ \bm{\alpha'}}$ and obtain locally optimal solutions $\bm{x}^*( \bm{\alpha'})$.

\subsection{CVAE model with GMM prior}

In \acronym{}, we use CVAE and GMM to build a locally optimal solution predictor.
For the trajectory optimization problem $\mathcal{P}_{\bm{\alpha}}$, this predictor can generate its solutions $\bm{x}^*(\bm{\alpha})$ with a clustering structure, conditioned on the problem parameter $\bm{\alpha}$.
Our work is based on the research that uses VAE and GMM to conduct clustering analysis on images \cite{jiang2016variational} and single-cell sequence \cite{xiong2019scale}.
We extend this previous work to a conditional generative framework \cite{Sohn2015learning} such that our model can generalize to problems with unseen parameter values.

Our proposed CVAE model takes the input data $\bm{x} \in \mathcal{U} \subseteq \R^n$ and the parameter $\bm{\alpha} \in \R^k$, as defined in Eq. \eqref{eq: parameterized optimization}.
The CVAE has a latent variable $\bm{z} \in \mathbb{R}^m$ with a prior distribution to be a GMM.
Here, we assume the GMM has $K$ components, and each component is a multivariate Gaussian with a diagonal covariance matrix.
We define a variable $c \in \mathbb{Z},  1 \leq c \leq K$, as a categorical variable that represents the component selected in the GMM.
For each Gaussian component selected by $c$, we define a pair of variables $\bm{\mu}_c \in \mathbb{R}^m$ and $\bm{\bm{\sigma}}_c^2 \in \mathbb{R}^m_{\geq 0}$, where $\bm{\mu}_c$ is the mean and $\bm{\sigma}_c^2$ is the diagonal element of the covariance.
We assume that both the variable $c$ and the corresponding $\bm{\mu}_c$, $\bm{\sigma}_c$ depend on the problem parameter $\bm{\alpha}$, such that the GMM prior can vary with respect to $\bm{\alpha}$.
\FloatBarrier
\begin{figure}[!htbp]
\centering
\begin{subfigure}{0.52\textwidth}
    \includegraphics[width=\textwidth]{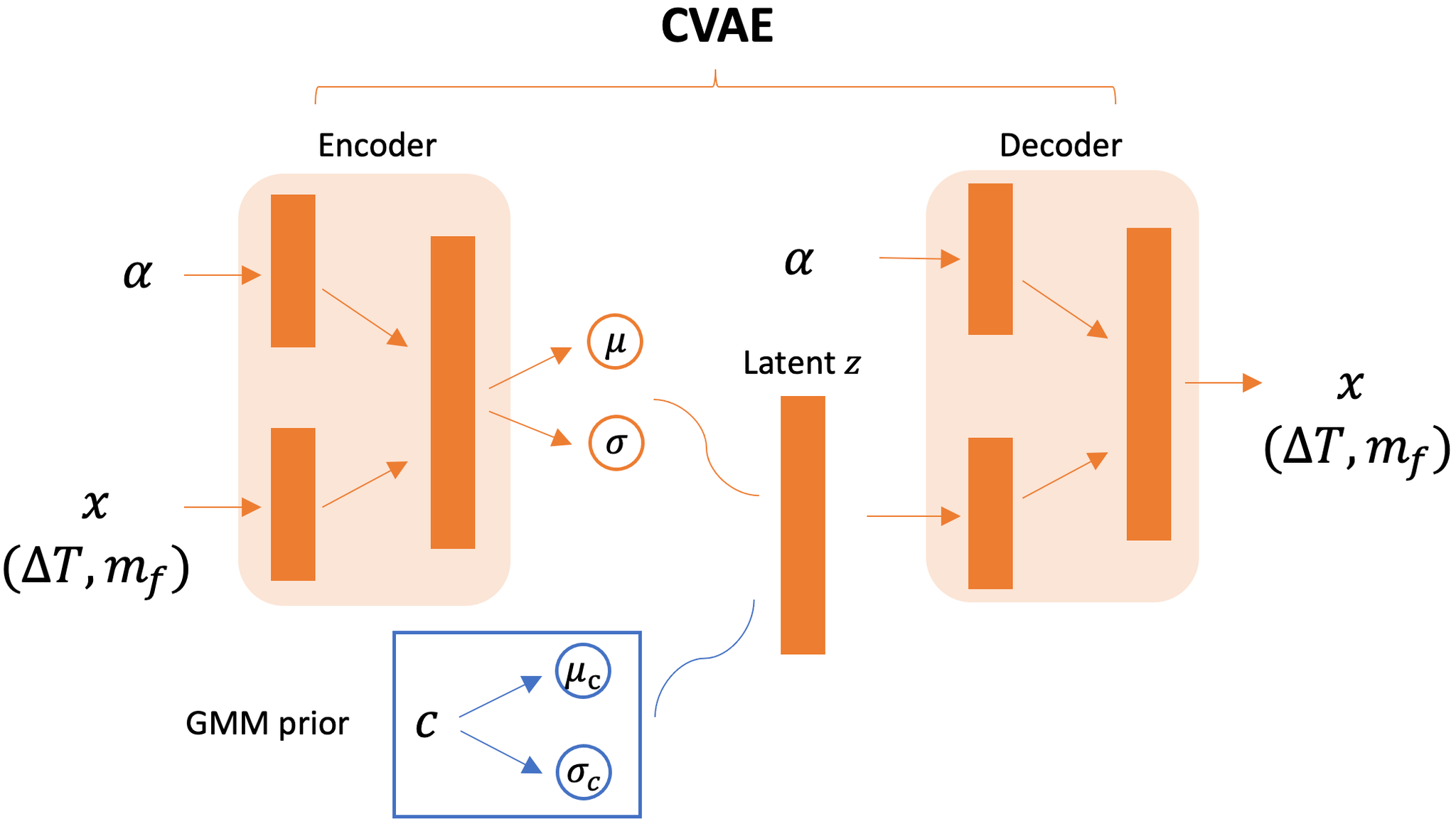}
    \caption{
    Our proposed CVAE model with GMM as the prior of the latent variable $\bm{z}$.}
    \label{fig:cvae architecture}
\end{subfigure}
\hfill
\begin{subfigure}{0.35\textwidth}
    \includegraphics[width=\textwidth]{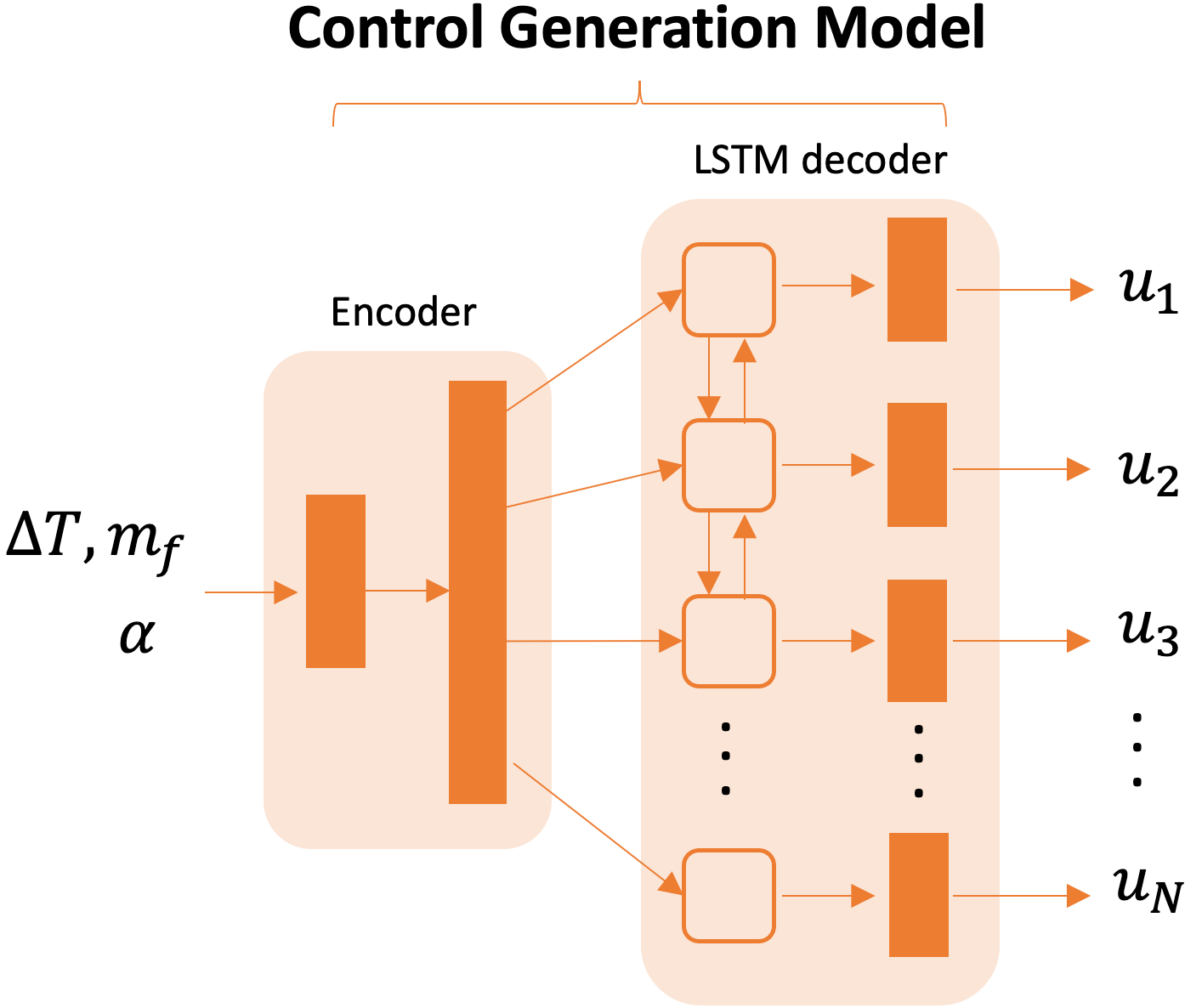} 
    \caption{Our proposed Control Generation Model using LSTM.} 
    \label{fig:control generation model architecture}
\end{subfigure}
\caption{(a): The CVAE model proposed in our \acronym{} framework to make structural predictions of the locally optimal solutions.
For De Jong's 5th function, we use the CVAE to predict the full-dimensional solution clusters.
For the transfer in CR3BP, we only use the CVAE to predict the time variables $\Delta T$ and final mass $m_f$ in the solutions, where the $\Delta T$ has hyperplane behaviors.
(b): The Control Generation Model we use to predict the corresponding control variable $\bm{u}_1, \bm{u}_2, ..., \bm{u}_N$ given samples of time $\Delta T$ and final mass $m_f$ in the CR3BP.
}
\label{Fig: model architecture}
\end{figure}
\FloatBarrier
In the CVAE, we model the input data $\bm{x}$, the parameter $\bm{\alpha}$, the latent variable $\bm{z}$, and $c$ as a conditional joint distribution $p(\bm{x}, c, \bm{z} | \bm{\alpha})$.
The generation process is defined in Eq.\eqref{eq: cvae generation} \cite{jiang2016variational}.
The categorical variable $c$ is drawn from a categorical distribution, denoted as $\text{Cat}(\boldsymbol{\pi}_{\bm{\alpha}})$, where \(\boldsymbol{\pi}_{\bm{\alpha}} = (\pi_{1,\bm{\alpha}}, \pi_{2,\bm{\alpha}}, ..., \pi_{K,\bm{\alpha}})\) represents the probability mass function of each component as functions of $\bm{\alpha}$. Each \(\pi_{i,\bm{\alpha}}\) is positive (\(\pi_{i,\bm{\alpha}} > 0\)), and for each $\bm{\alpha}$ the sum of all \(\pi_{i,\bm{\alpha}}\) values is 1 (\(\sum_{i=1}^{K} \pi_{i,\bm{\alpha}} = 1\)), ensuring that \(\boldsymbol{\pi}_{\bm{\alpha}}\) is a valid probability distribution.
Given the sampled $c$, the latent variable $\bm{z}$ is sampled from a Gaussian distribution $\mathcal{N}(\bm{\mu}_{c,\bm{\alpha}}, \bm{\sigma}^2_{c,\bm{\alpha}} \mathbf{I})$ parameterized by the corresponding $\bm{\mu}_{c,\bm{\alpha}}, \bm{\sigma}^2_{c,\bm{\alpha}}$.
Finally, the output $\bm{x}$ can be reconstructed from the latent variable $\bm{z}$ with $p(\bm{x}|\bm{z}, \bm{\alpha})$ which can be approximated by the decoder of the CVAE.
\begin{align}\label{eq: cvae generation}
    c \sim \text{Cat}(\boldsymbol{\pi}_{\bm{\alpha}}), \quad \bm{z} \sim \mathcal{N}(\bm{\mu}_{c,\bm{\alpha}}, \bm{\sigma}^2_{c,\bm{\alpha}}\mathbf{I} ), \quad \bm{x} \sim p(\bm{x}|\bm{z}, \bm{\alpha})
\end{align}

According to this generation process, where $\bm{x}$ and $c$ is independent conditioned on $\bm{z}$, we can factorize $p(\bm{x}, c, \bm{z} | \bm{\alpha}) $ as $ p(\bm{x}|\bm{z}, \bm{\alpha})p(\bm{z}|c, \bm{\alpha})p(c | \bm{\alpha})$.
$p(c | \bm{\alpha}) = \text{Cat}(c; \bm{\pi}_{\bm{\alpha}})
$ is a categorical distribution.
$p(\bm{z}|c, \bm{\alpha}) = \mathcal{N}(z; \bm{\mu}_{c,\bm{\alpha}}, \bm{\sigma}^2_{c,\bm{\alpha}}\mathbf{I} )$ is modeled as a multivariate diagonal Gaussian.
$\bm{\pi}_{\bm{\alpha}}, \bm{\mu}_{c,\bm{\alpha}}$ and $\bm{\sigma}^2_{c,\bm{\alpha}}$ are functions of parameter $\bm{\alpha}$ and can be parameterized by neural netowrks with parameter $\theta$.
$p(\bm{x}|\bm{z}, \bm{\alpha})$ is approximated with a decoder $p_{\psi}(\bm{x}|\bm{z}, \bm{\alpha})$, parameterized by the neural network with parameters $\psi$.

Using an encoder-decoder architecture, the CVAE first maps the original data $\bm{x}$ to the latent variable $c$ and $\bm{z}$ with an encoder $q_{\phi}(\bm{z}, c|\bm{x}, \bm{\alpha})$, parameterized by the neural network with parameters $\phi$.
In the latent space, we sample the variable $c$ and $\bm{z}$ as described in Eq. \eqref{eq: cvae generation}.
The latent variable $\bm{z}$ is supposed to lie on the mixture-of-Gaussian manifold defined by $\bm{\pi_{\alpha}}$, $\bm{\mu}_{c, \bm{\alpha}}$ and $\bm{\sigma}^2_{c, \bm{\alpha}}$.
Next, through the decoder $p_{\psi}(\bm{x}|\bm{z}, \bm{\alpha})$, we can reconstruct the original data $\bm{x}$ from samples of $\bm{z}$ and the parameter $\bm{\alpha}$.

The objective of this CVAE is to maximize the conditional log-likelihood $\log p(\bm{x}|\bm{\alpha})$ of the output samples $\bm{x}$.
However, directly maximizing the conditional log-likelihood is mathematically intractable.
Instead, we optimize the Evidence Lower Bound (ELBO) as the loss function which is the lower bound of the conditional log-likelihood obtained by Jensen's inequality \cite{Sohn2015learning, jiang2016variational}.
\begin{align}
    \log p_{\psi}(\bm{x}|\bm{\alpha}) &= \log \int_{\bm{z}} \sum_c p_{\psi}(\bm{x}, \bm{z}, c | \bm{\alpha}) d\bm{z} \nonumber \\
    &\geq \mathbb{E}_{\bm{z}, c \sim q_{\phi}(\bm{z}, c|\bm{x}, \bm{\alpha})} \left[ \log \frac{p_{\psi}(\bm{x}, \bm{z}, c |\bm{\alpha})}{q_{\phi}(\bm{z}, c|\bm{x}, \bm{\alpha})} \right]  \nonumber \\
    &= \mathbb{E}_{\bm{z}, c \sim q_{\phi}(\bm{z}, c|\bm{x}, \bm{\alpha})} \left[\log p_{\psi}(\bm{x}, \bm{z}, c |\bm{\alpha}) - \log q_{\phi}(\bm{z}, c|\bm{x}, \bm{\alpha})\right] \nonumber \\
    &= \mathbb{E}_{\bm{z}, c \sim q_{\phi}(\bm{z}, c|\bm{x}, \bm{\alpha})} [\log p_{\psi}(\bm{x}|\bm{z}, \bm{\alpha}) + \log p(\bm{z}|c, \bm{\alpha}) + \log p(c | \bm{\alpha}) \nonumber \\ 
    &\quad   -\log q_{\phi}(\bm{z}, c|\bm{x}, \bm{\alpha})]  \label{eq: elbo}
\end{align}

To better understand the intuition behind the ELBO in the last line of Eq. \eqref{eq: elbo}, we can rewrite the ELBO as the following objective function:
\begin{align} \label{eq: cvae objective}
     \mathcal{L}_{CVAE} := \mathbb{E}_{\bm{z}, c \sim q_{\phi}(\bm{z}, c| \bm{x}, \bm{\alpha})} [\log p_{\psi}(\bm{x} | \bm{z}, \bm{\alpha})] - D_{KL}(q_{\phi}(\bm{z}, c|\bm{x}, \bm{\alpha}) || p(\bm{z}, c | \bm{\alpha}))
\end{align}
The first part of the loss function in Eq. \eqref{eq: cvae objective} intends to reduce the reconstruction error of the sample $\bm{x}$ compared to the input data.
Here we use a mean squared error (MSE) loss to measure the error between the reconstructed $\bm{x}$ and the input $\bm{x}$.
The second part of the loss function in Eq. \eqref{eq: cvae objective} is a Kullback–Leibler (KL) divergence term that enforces the posterior distribution of the latent $\bm{z}$ approximated by the encoder to lie on a mixture-of-Gaussian manifold.
To make sure that backpropagation works through the random samples of latent variable $\bm{z}$, we use the re-parameterization trick when computing the KL divergence term.
We refer to \cite{jiang2016variational} for more details on computing the KL divergence term for VAE with GMM prior and refer to \cite{Sohn2015learning} for the detailed extension from VAE to CVAE.

In our proposed CVAE, we use Multilayer Perceptrons (MLP) with Leaky Rectified Linear Unit (LeakyReLU) as the activation function in the hidden layers and use Sigmoid function as the activation function before the final output $x$.
The CVAE architecture is depicted in Fig. \ref{fig:cvae architecture}.
For the encoder, we first embed the input $\bm{x}$ and $\bm{\alpha}$ separately.
Then we concatenate this embedding and encode them to the posterior mean $\bm{\mu}$ and the diagonal of the covariance $\bm{\sigma}^2$ of the latent variable $\bm{z}$.
We sample the latent variable $\bm{z}$ using the reparameterization trick \cite{kingma2013auto}.
For the decoder, we first embed the sampled latent variable $\bm{z}$ and use the same layer as the encoder to embed $\bm{\alpha}$. 
Finally, we concatenate the embeddings and decode to the reconstructed $\bm{x}$.
We also construct the GMM variable $c$, the mean $\bm{\mu}_c$ and covariance $\bm{\sigma}_c^2$ for each GMM component as functions of $\bm{\alpha}$, and keep them in the training loop.
In the training time, we sample $\bm{z}$ from the posterior $\bm{\mu}, \bm{\sigma}^2$ approximated by the encoder and enforce it to lie on a mixture-of-Gaussian manifold.
During the testing time, we can sample $c, \bm{z}$ from the updated GMM prior $\bm{\pi}_c$ , $\bm{\mu}_c$ and $\bm{\sigma}_c$ and then generate $\bm{x}$ through the decoder conditioned on $\bm{\alpha}$.

\section{Experiments} \label{sec: expriment}


We consider two global search problems to evaluate the efficacy of our proposed \acronym{} framework.
In the first test problem, we aim to search all the locally optimal solutions of De Jong's 5th function, a commonly used global optimization test function.
In this two-dimensional problem, we are able to manually set up where the local optima $\bm{x}^*(\alpha)$  are and how they will change with respect to the parameter $\alpha$.
We will demonstrate the \acronym{}'s ability to predict locally optimal solutions when an unseen parameter value is given.
The second problem is a minimum-fuel low-thrust transfer in the CR3BP.
Our goal is to conduct a more efficient global search of this trajectory optimization problem to find qualitatively different locally optimal solutions.
For this example, we propose a control generation model together with the aforementioned CVAE to make predictions of the solutions.
\FloatBarrier
\begin{figure}[!htbp]
    \centering
    \includegraphics[width=0.3\linewidth]{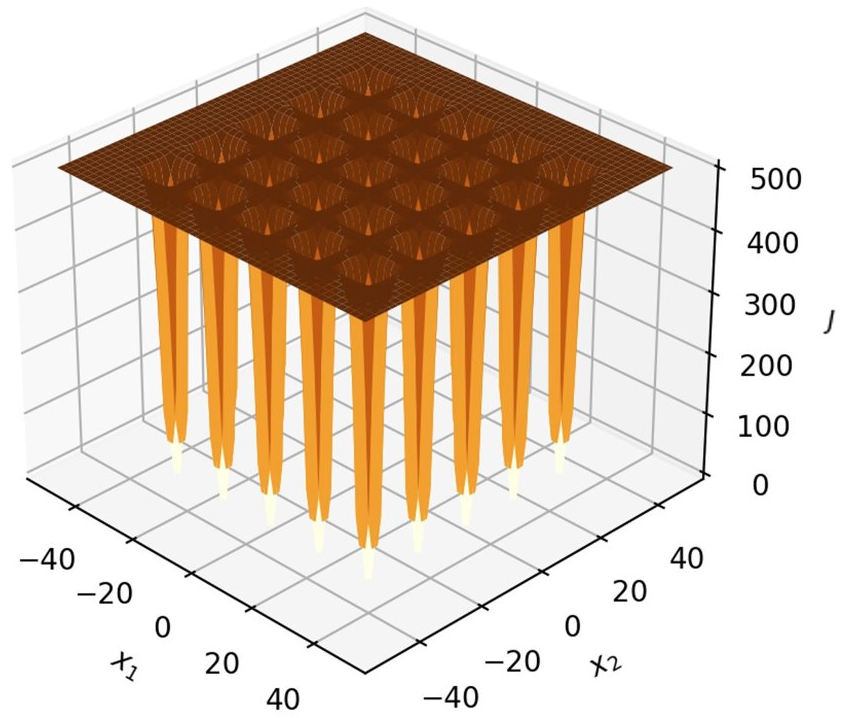}
    \caption{An example of a 2-dimensional De Jong's 5th problem with 25 local minima.}
    \label{Fig: de jong 5}
\end{figure}
\FloatBarrier
In this paper, we benchmark our method to MBH, the state-of-the-art global optimization algorithm.
MBH is a two-level algorithm.
Globally the MBH iteratively samples an initialization $\bm{x}_0$ with a fixed distribution, such as the uniform \cite{englander2017automated}, Gaussian, Cauchy, or Pareto distributions \cite{englander2014tuning}.
When it finds a feasible solution, it will run a local search through some random perturbation of the initialization.
In the following experiment, we will compare our proposed method with the global level of the MBH method, showing that global search will achieve higher efficiency when initialized from our CVAE prediction rather than the initial guesses drawn from any fixed global distribution.

\subsection{De Jong's 5th Function}

The parameterized optimization problem  $\mathcal{P}_{\alpha}$ to minimize De Jong’s 5th function is defined as follows:
\begin{align} \label{eq: de jong 5}
    &\min \quad J(\bm{x}) = \biggl( 0.002  + \sum^n_{i = 1}\frac{1}{1 + (\bm{x}_1 - \bar a_{1i})^6 + (\bm{x}_2 - \bar a_{2i})^6}  \biggr)^{-1}  \nonumber \\
    & 
    \text{where} \quad \bar a =
    \begin{pmatrix}
        \cos \alpha & - \sin \alpha \\
        \sin \alpha & \cos \alpha
    \end{pmatrix}
    a, \quad 
    a = 
    \begin{pmatrix}
        a_{11} & a_{12} & a_{13} &... &a_{1n}\\
        a_{21} & a_{22} & a_{23} &... &a_{2n}
    \end{pmatrix} 
\end{align}
where $\bm{x} = (x_1, x_2) \in \mathbb{R}^2, x_1, x_2 \in [-50, 50]$ and each column of $\bar a$ defines a locally minimal solution of the function $J(\bm{x})$.
Here, $\alpha \in [0, \pi / 2]$ is a one-dimensional problem parameter that controls the rotation of the locally minimal solutions to the original De Jong's 5th function.
When $\alpha = 0$, the locally minimal solutions are the columns of the matrix $a$.
An example of this De Jong's 5th function with 25 local minima is depicted in Figure \ref{Fig: de jong 5}.
\FloatBarrier
\begin{table}[ht]
\centering
\small
\begin{tabular}{p{2.2cm}p{2.2cm}p{2.8cm}p{2.2cm}}
\hline
\multicolumn{4}{c}{\textbf{CVAE}} \\
\hline
\textbf{Layer Name} & \textbf{Layer Size} & \textbf{Layer Name} & \textbf{Layer Size} \\
\hline
Embed\_$\bm{x}$\_layer & [2, 32, 64, 64] & Embed\_$\alpha$\_layer & [1, 32, 64, 64] \\
Encode\_layer & [128, 64, 64] & Encode\_$\bm{\mu}$/$\bm{\sigma}^2$\_layer & [64, 32, 2] \\
Embed\_$\bm{z}$\_layer & [2, 32, 64, 64] & Decode\_$\bm{x}$\_layer & [128, 64, 64, 2] \\
\hline
\end{tabular}
\caption{CVAE architecture for testing the De Jong's 5th function.}
\label{table: CVAE dejong 5}
\vspace{-5pt}
\end{table}
\FloatBarrier
\begin{figure}[!htbp]
    \centering
    \begin{subfigure}[b]{1.0\textwidth}
        \includegraphics[width=\textwidth]{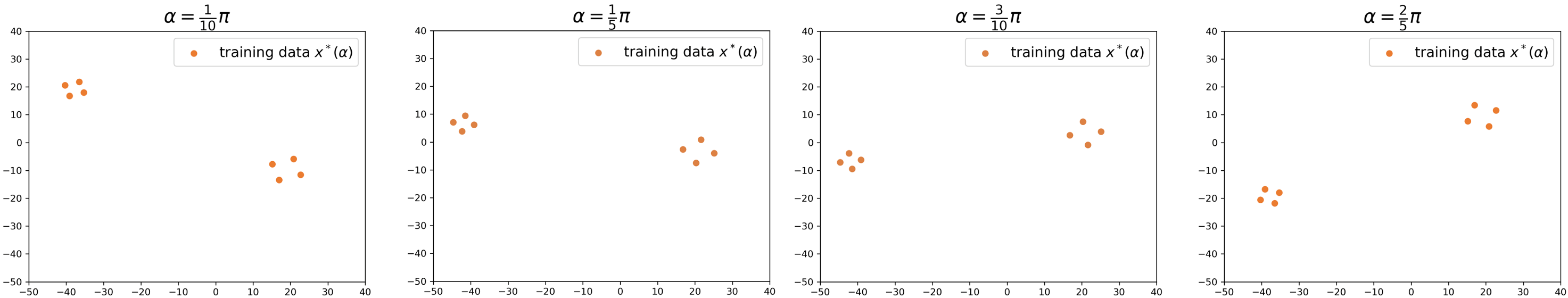}
        \caption{The training data for De Jong's 5th function with 2 clusters of locally optimal solutions $\bm{x}^*(\alpha)$, where $\alpha$ is drawn from a grid in $[0, \pi / 2]$.}
        \label{Fig: de jong 5 training data}
    \end{subfigure}
    \begin{subfigure}[b]{1.0\textwidth}
        \includegraphics[width=\textwidth]{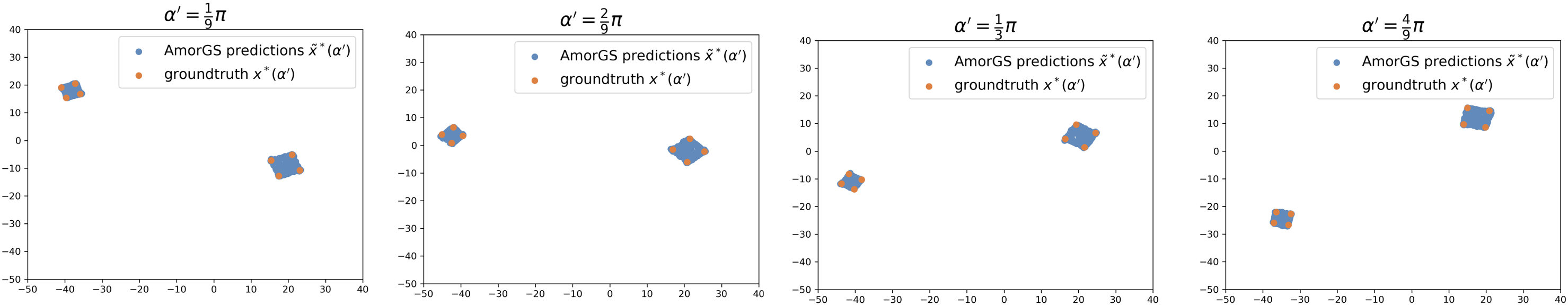}
        \caption{\acronym{} predictions of 1k locally optimal solutions $\bm{\tilde x}^*(\alpha')$ with $\alpha' = \frac{1}{9} \pi, \frac{2}{9} \pi, \frac{1}{3} \pi, \frac{4}{9} \pi$ to De Jong's 5th function.
    The \textcolor{blue}{blue} dots are predictions $\bm{\tilde x}^*(\alpha')$ from \acronym{} and the \textcolor{DarkOrange}{orange} dots are the corresponding groundtruth $\bm{x}^*(\alpha')$.
    \acronym{} can not only accurately predict the locations of the locally optimal solutions $\bm{x}^*(\alpha')$ but also the rotated shapes of each cluster for each unseen $\alpha'$.}
        \label{Fig: de jong 5 test result}
    \end{subfigure}
    \caption{Training data and testing results for De Jong's 5th function.}
    \label{Fig: de jong 5 result}
\end{figure}
\FloatBarrier
For this problem, we aim to test the performance of our \acronym{} framework to predict the locally optimal solutions to the problem $\mathcal{P}_{\alpha'}$ when \textit{a priori} unknown rotation $\alpha'$ is presented.
To mimic the clustering pattern of the trajectory optimization problem, we set the matrix $a$ as follows:
\begin{align} \label{eq: a matrix}
    a = \begin{pmatrix}
        -32 & -32 & -28 & -28 & 12 & 12 & 18 & 18\\
        32 & 28 & 32 & 28 & - 12 & - 18 & - 12 & - 18
    \end{pmatrix}
\end{align}
such that the eight local optimal solutions are grouped into 2 clusters, each with 4 solutions.
During the data preparation process, we solve this De Jong's 5th function using the BFGS algorithm, initialized from the grid in $[-50, 50] \times [-50, 50] $.
We collect 100k locally optimal solution $\{\bm{x}^*_i(\alpha) \}_{i=1}^{100k}$ for 51 distinct values of $\alpha$ that is drawn from a grid in $[0, \pi / 2]$.
This is to make our CVAE model fully aware of how the distribution of $\bm{x}^*(\alpha)$ changes with respect to $\alpha$.
We visualize part of our constructed dataset in Fig. \ref{Fig: de jong 5 training data}.

In this example, we build our CVAE with PyTorch \cite{paszke2017automatic} using the architecture specified in Table \ref{table: CVAE dejong 5}.
The GMM prior we use for the CVAE consists of two components and each component is a 2-dimensional Gaussian with diagonal covariance matrix.
Considering that the density of the solution clusters remains unchanged for this particular example, we assume that the GMM prior is independent of the parameter $\alpha$ .
We demonstrate that by solely modulating the conditional input $\alpha$ of the decoder, we can successfully reconstruct the solution clusters for unseen $\alpha$.

We test our \acronym{} framework to predict the locally minimum solutions $\bm{x}^*(\alpha')$ to De Jong's 5th function with \textit{a priori} unseen $\alpha'$.
We make 1000 predictions for each test $\alpha' = \frac{1}{9} \pi, \frac{2}{9} \pi, \frac{1}{3} \pi, \frac{4}{9} \pi$, which are not seen in the training dataset.
Given the analytical form in Eq. \eqref{eq: de jong 5}, matrix $a$ set up as in Eq. \eqref{eq: a matrix}, and the rotation $\alpha'$, we can generate the groundtruth as a comparison to the \acronym{} results.
In Fig. \ref{Fig: de jong 5 test result}, the orange dots are the groundtruth of the locally optimal solution $\bm{x}^*(\alpha')$ given current orientation $\alpha'$, and the blue dots are the predictions from \acronym{}.
We can see that the \acronym{} predictions $\bm{\tilde x}^*(\alpha')$ closely approximate the locations of the solution clusters $\bm{x}^*(\alpha')$, even with \textit{a priori} unknown rotation $\alpha'$.
It also generates the correct shape of the clusters after the rotation.
Although there might be outliers in the prediction of \acronym{}, the likelihood of this occurrence is extremely low, approximately one in a thousand.
This result demonstrates the generalizability of our \acronym{} method to predict solutions to a family of problems in Eq. \eqref{eq: parameterized optimization} with previous unseen parameter values $\alpha'$.

\subsection{The Minimum-fuel low-thrust transfer in the CR3BP}

\textbf{The CR3BP}

In this study, the dynamical model used to describe the spacecraft dynamics is the CR3BP, which describes the motion of a point-mass in a three-body system e.g., a spacecraft orbiting the Earth-Moon system.
The CR3BP offers a \textit{good} first-order approximation of complex dynamical systems and is a valuable toolset for analysis.
The CR3BP describes the motion of a spacecraft, whose mass is assumed to be negligible, under the gravitational influence of two celestial bodies, such as the Earth and the Moon, which rotate about their common center of mass in circular orbits.
To simplify the analysis, we write the equations of motion of the spacecraft in a synodic reference frame rotating at the same rate as the two primaries.
The analysis can be further simplified by non-dimensionalizing the equations using a suitable choice of units which reduces the number of parameters in the problem to one, namely, the mass parameter $\mu = m_2 / (m_1 + m_2)$, where $m_1$ is the mass of the primary and $m_2 \leq m_1$ is the mass of the secondary.
With this choice of units, the gravitational constant and the mean motion both become unity, leading to the following equations of motion:
\begin{align}
    \ddot{x} - 2\dot{y} = -\bar{U}_x, \quad
    \ddot{y} + 2\dot{x} = -\bar{U}_y, \quad
    \ddot{z} = -\bar{U}_z, \nonumber
\end{align}
where $\bar{U}(r_1(x,y,z),r_2(x,y,z)) = -\frac{1}{2}\left((1-\mu) r_1^2 + \mu r_2^2\right) - \frac{1-\mu}{r_1} - \frac{\mu}{r_2}$, and the distance $r_1 = \sqrt{(x + \mu)^2 + y^2 + z^2)}$, and $r_2 = \sqrt{(x - (1-\mu))^2 + y^2 + z^2)}$.
$\bar{U}(r_1(x,y,z),r_2(x,y,z))$ is the \textit{effective} gravitational potential; $r_1$ and $r_2$ are the distances between the spacecraft to the primary and the secondary, respectively, in the rotating frame coordinates.
In this dynamical model, there exist five equilibrium points known as the Lagrange points, around which it is possible to obtain periodic orbits such as Lyapunov, Halo, or resonant orbits.

\textbf{Minimum-fuel low-thrust transfer problem}

We consider a minimum-fuel transfer of a low-thrust spacecraft as our parameterized optimization problem $\mathcal{P}_{\alpha}$.
The spacecraft starts from a Geostationary Transfer Orbit spiral and reaches a stable manifold arc of a Lyapunov orbit around the L1 Lagrange point.
In this experiment, we fix the constant specific impulse (CSI) $I_{sp} = 1000$ s, initial fuel mass to be $700$ kg, and dry mass to be $300$ kg.
A one-dimensional problem parameter $\alpha$ we want to test is the maximum allowable thrust of the spacecraft.
We set the bound for $\alpha \in [0.1, 1]\text{N}$, thus the initial mass acceleration of the spacecraft will range from $10^{-4} \ \text{m}/\text{s}^2$ to $10^{-3} \ \text{m}/\text{s}^2$.
Our goal is to conduct an efficient global search for the locally optimal trajectory solutions when \textit{a priori} unseen maximum thrust $\alpha$ is provided.
\FloatBarrier
\begin{figure}[!htbp]
    \centering
    \includegraphics[width=0.95\linewidth]{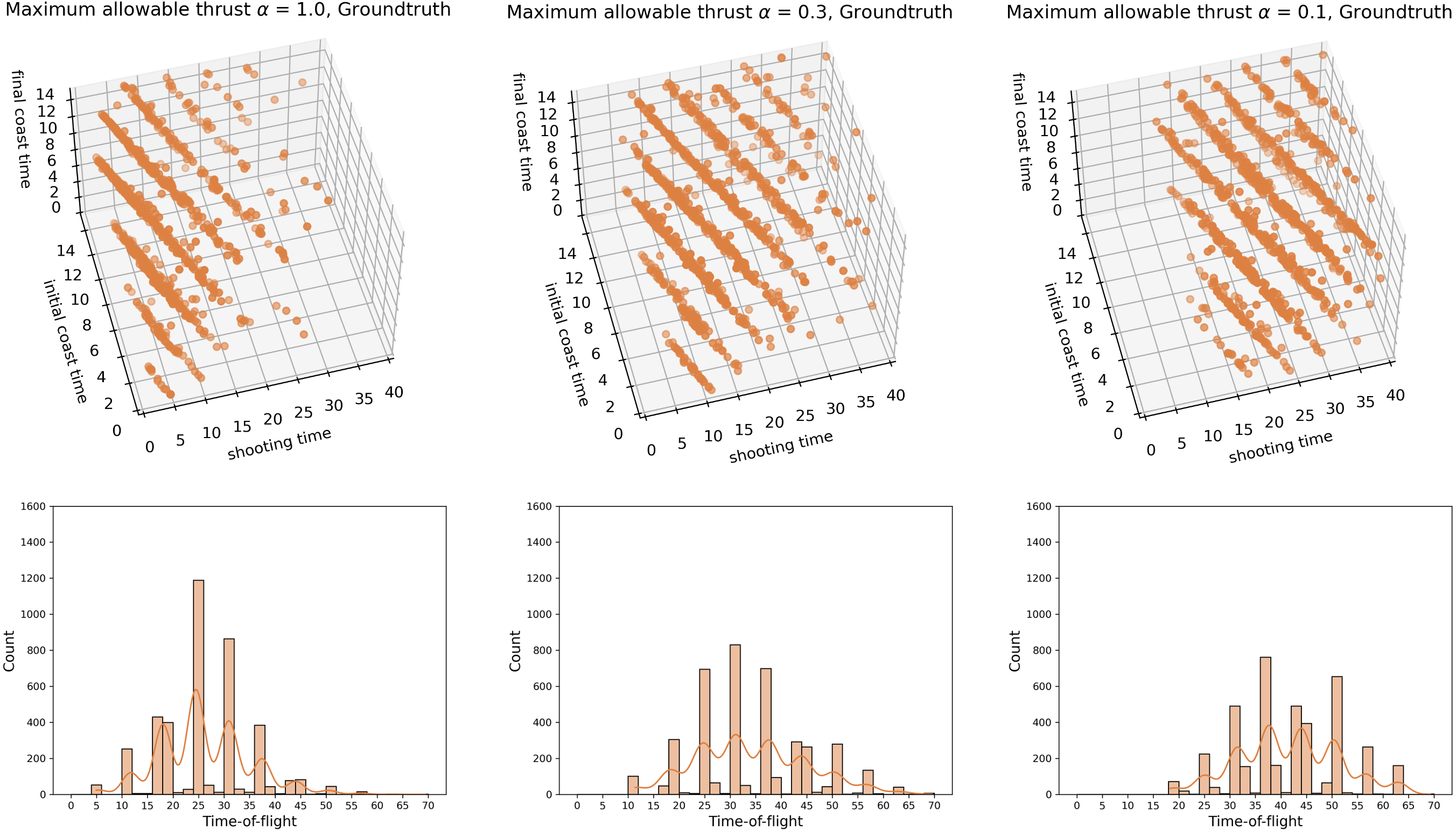}
    \caption{The hyperplane behaviors of locally optimal solution $x^*(\alpha)$ change with respect to three different maximum allowable thrust   $\alpha$ values.
    As we decrease the maximum allowable thrust $\alpha$, the hyperplanes will move toward the area that has a longer time-of-flight.
    The histograms of time-of-flight also show the distribution shift for the solutions as we change the $\alpha$.}
    \label{Fig: cr3bp different thrust}
\end{figure}
\FloatBarrier
We use a forward-backward shooting transcription to formulate a nonlinear program $\mathcal{P}_{\alpha}$ as in Eq. \eqref{eq: parameterized optimization} with the variable $\bm{x} \in \mathbb{R}^{3N+4}$ and the parameter $\alpha \in [0.1, 1.0]$ representing the maximum allowable thrust.
With this forward-backward shooting transcription, the variable $\bm{x}$ to optimize is defined as follows:
\begin{align} \label{eq: cr3bp variable}
    \bm{x} = (\Delta T_{shooting}, \Delta T_{coast}^{initial}, \Delta T_{coast}^{final}, m_f, \bm{u}_1, \bm{u}_2, ..., \bm{u}_N)
\end{align}
In this $(3N + 4)$ -dimensional variable $\bm{x}$, $N$ is the number of discretized control segments.
In a finite-burn LT model, both the magnitude and direction of the thrust remain constant across each segment.
In each segment, there is a variable $\bm{u}_i \in \mathcal{U} \subseteq \R^3, i = 1, 2, ..., N$, with a three-dimensional degree of freedom, representing the direction and magnitude of the thrust vector.
The remaining four variables are the time variables $\Delta T$ including the initial coast time $\Delta T_{coast}^{initial}$, the final coast time $\Delta T_{coast}^{final}$, the shooting time $\Delta T_{shooting}$,  and the final mass $m_f$ at the terminal boundary.

In this problem $\mathcal{P}_{\alpha}$, the objective function $J$ is set to maximize the final mass $m_f$, thereby seeking solutions that minimize fuel consumption.
The only constraint $c$ in this problem is an equality constraint that enforces the system state at the midpoint to match when integrated forwardly and backwardly based on the dynamical model.
This problem is solved using pydylan, a python interface of the Dynamically Leveraged Automated (N) Multibody Trajectory Optimization (DyLAN) \cite{beeson_dylan_2022} solver and SNOPT \cite{gill_snopt_2005}.

For the training data, we collect 300k locally optimal \textit{good} solutions of the problem $\mathcal{P}_{\alpha}$, whose final mass $m_f \geq 415$kg.
The problem $\mathcal{P}_{\alpha}$ is solved with 12 levels of maximum allowable thrust $\alpha = 0.1, 0.13, 0.16, 0.2, 0.3, 0.4, 0.5, 0.6, 0.7, 0.8, 0.9, 1.0$.
We select $\alpha$ samples more densely in the lower range, as the solution structure experiences more significant changes within this range.
The initializations are uniformly sampled: $\Delta T_{shooting} \in [0, 40]\text{TU}, \Delta T_{coast}^{initial}, \Delta T_{coast}^{final} \in [0, 15]\text{TU}$, and $ m_f \in [350, 450]\text{kg}$.
The direction of control $\bm{u}$ is uniformly sampled from $[0, 2\pi]$ and its magnitude is also uniformly sampled from $[0, 1]$.
To thoroughly exploit the solution space, we set the maximum running time for the solver $\pi$ to be 500s.

\textbf{Hyperplane structure in the solutions}

We observe that the three-dimensional time variables $\Delta T_{shooting}, \Delta T_{coast}^{initial}, \Delta T_{coast}^{final}$ from the locally optimal solutions in the training data fall into several parallel hyperplanes.
Figure \ref{Fig: cr3bp different thrust} displays this hyperplane pattern for 1k local optimal solutions $\bm{x}^*(\alpha)$ with $\alpha = 1.0, 0.3, 0.1$.
Moreover, the sum of these three time variables $\Delta T_{shooting}, \Delta T_{coast}^{initial}, \Delta T_{coast}^{final}$, which corresponds to the time-of-flight, remains approximately constant within each plane.
As we move from the lower to the upper hyperplanes, the trajectory solutions increasingly exhibit higher time-of-flight values.
The second row of Figure \ref{Fig: cr3bp different thrust} presents a histogram of the time-of-flight for the solutions in the first row. 
Each peak in the histogram represents a hyperplane shown in the adjacent plots.
This result supports the hypothesis that the local optimal solutions of the trajectory optimization problem $\mathcal{P}_{\alpha}$ tend to be grouped into clusters \cite{yam2011low, englander2012automated}.
\FloatBarrier
\begin{figure}[!htbp]
    \centering
    \includegraphics[width=0.9\linewidth]{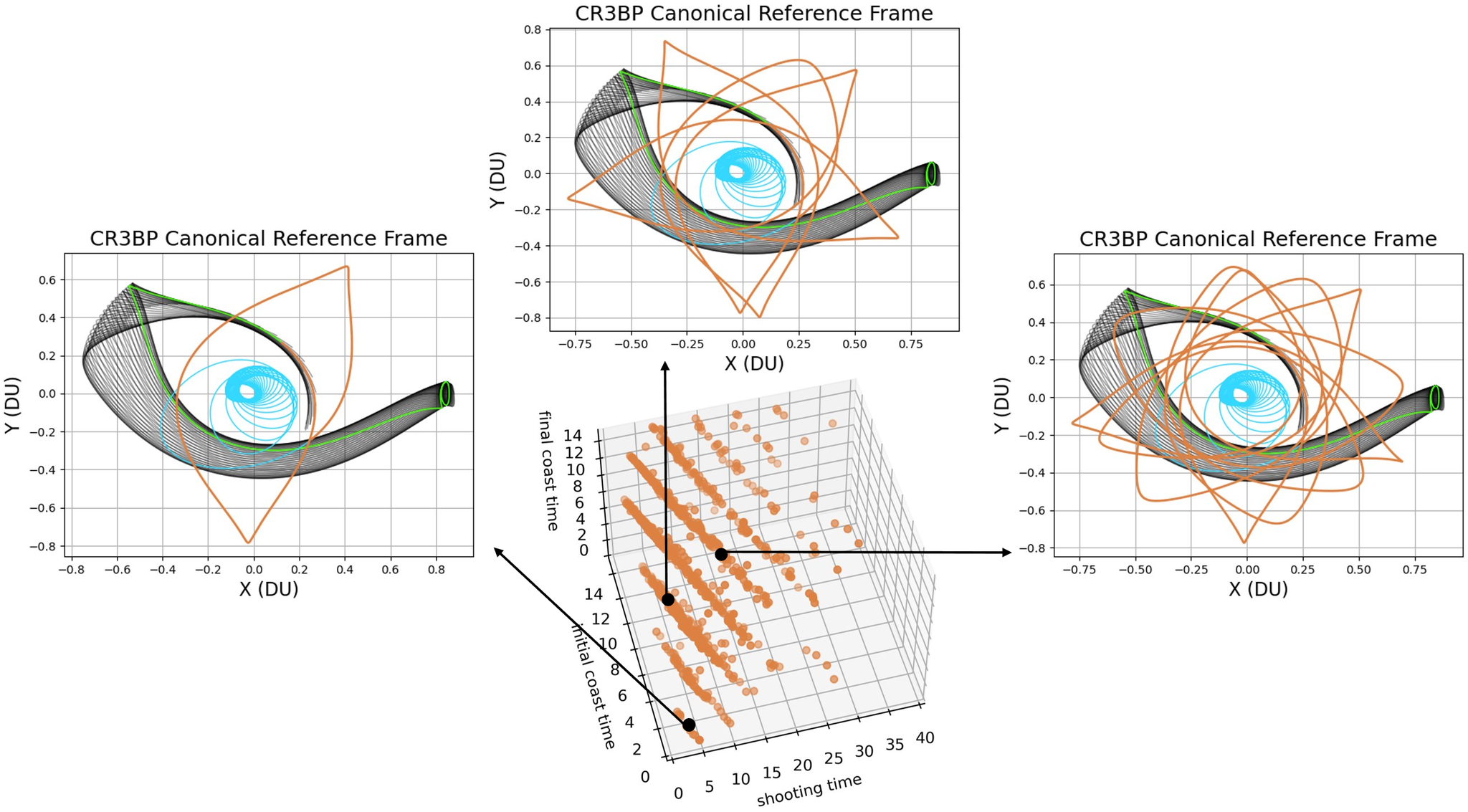}
    \caption{
    Three qualitatively different locally optimal trajectory solutions $\bm{x}^*(\alpha)$ to a minimum-fuel low-thrust transfer in the CR3BP $\mathcal{P}_{\alpha}$ are sampled from three different hyperplanes of time variables,  with the maximum allowable thrust $\alpha = 1.0$.
    The distance units are the Earth-Moon CR3BP distance units i.e., 1 DU = 384, 400 km.
    These trajectories (in \textcolor{DarkOrange}{orange}) start from the end of an LT spiral (in \textcolor{cyan}{cyan}) to a target arc (in \textcolor{Chartreuse}{green}) on the stable invariant manifold (in black) of a Lyapunov orbit around $\mathcal{L}_1$ Lagrange point.
    From the bottom left to the top right hyperplane, the time-of-flight increases because the trajectories contain more revolutions.
    }
    \label{Fig: trajectory}
\end{figure}
\FloatBarrier
The locally optimal solutions $x^*(\alpha)$ generally exhibit similar hyperplane behaviors across various maximum allowable thrust $\alpha$, but a closer inspection reveals that the hyperplanes are indeed influenced by the specific value of $\alpha$.
When we compare hyperplanes and histograms across $\alpha = 1.0, 0.3, 0.1$ in Figure \ref{Fig: cr3bp different thrust}, we see that a higher maximum allowable thrust $\alpha = 1.0$ is generally associated with locally optimal solutions $x^*(\alpha)$ with shorter time-of-flight.
On the contrary, for a significantly lower maximum allowable thrust $\alpha = 0.1$, the solutions with a short time-of-flight become infeasible.
Thus, as the maximum allowable thrust $\alpha$ decreases, the hyperplanes are likely to move toward an area with a larger time-of-flight.
\FloatBarrier
\begin{table}[!htbp]
\centering
\footnotesize
\begin{tabular}{p{2.5cm}p{3.6cm}p{2.5cm}p{3cm}}
\hline
\multicolumn{4}{c}{\textbf{CVAE}} \\
\hline
\textbf{Layer Name} & \textbf{Layer Size} & \textbf{Layer Name} & \textbf{Layer Size} \\
\hline
Embed\_$\bm{x}$\_layer & [4, 1024, 1024, 1024, 1024] & Embed\_$\alpha$\_layer & [1, 256, 256, 256, 256] \\
Encode\_layer & [1280, 512, 512, 512, 128] & Encode\_$\bm{\mu}$/$\bm{\sigma}^2$\_layer & [128, 128, 128, 4] \\
Embed\_$\bm{z}$\_layer & [4, 1024, 1024, 1024, 1024] & Decode\_$\bm{x}$\_layer & [1280, 512, 512, 512, 4] \\
\hline
\multicolumn{4}{c}{\textbf{Control Generation Model}} \\
\hline
\textbf{Layer Name} & \textbf{Layer Size} & \textbf{Layer Name} & \textbf{Layer Size} \\
\hline
Encoder\_layer & [5, 512, 512, 512] & Decoder\_layer & [512, 512, 512, 3] \\
\hline
\end{tabular}
\caption{The CVAE and Control Generation Network architecture for minimum-fuel low-thrust transfer in the CR3BP.}
\label{table: CVAE architecture cr3bp}
\end{table}
\FloatBarrier
\FloatBarrier
\begin{figure}[!htbp]
    \centering
    \includegraphics[width=1.0\linewidth]{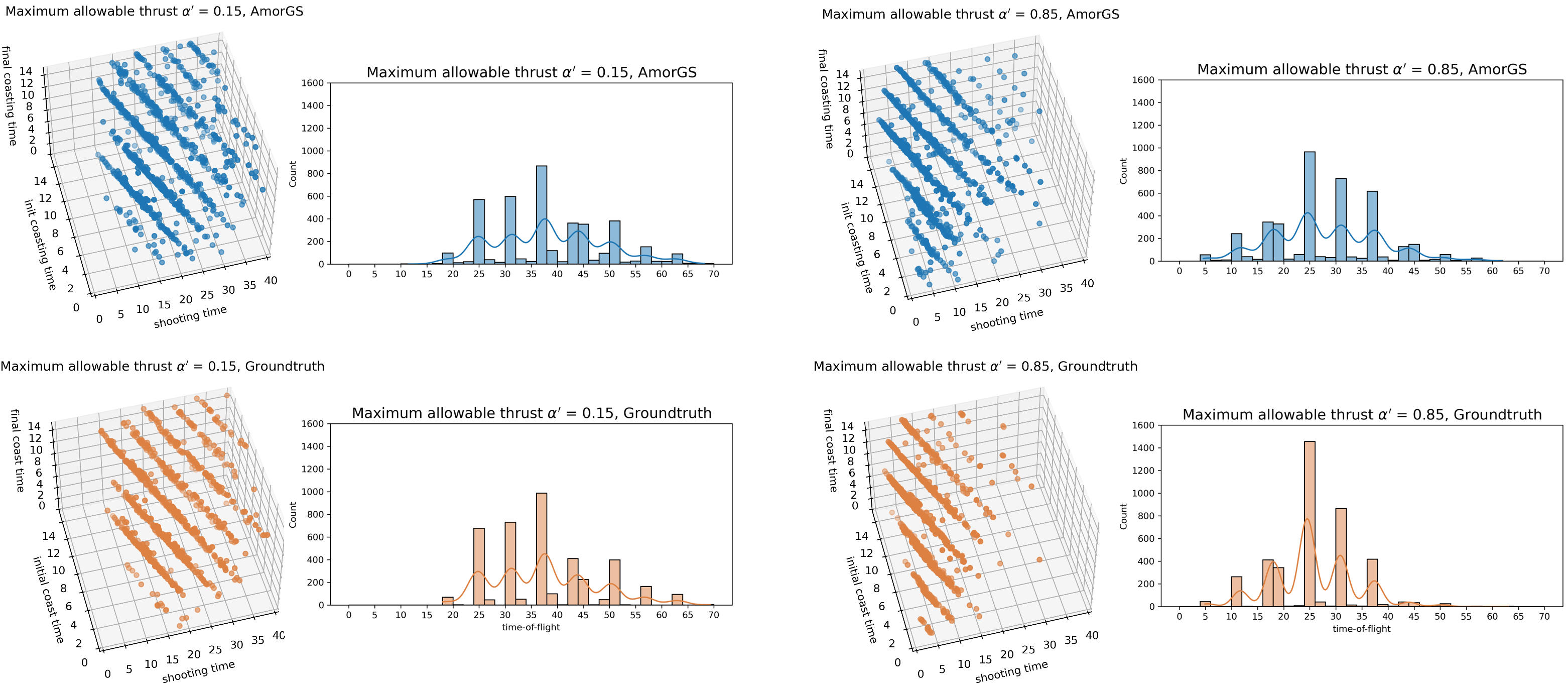}
    \caption{The comparison of hyperplane solution structure and corresponding histogram of time-of-flight between \acronym{} predictions and groundtruth for a minimum-fuel low-thrust transfer in the CR3BP.
    The prediction plots are in \textcolor{blue}{blue} and the groundtruths are in \textcolor{orange}{organge}.
    The maximum allowable thrust $\alpha'$ to test are $0.15$ (Left) and $0.85$ (Right) which are \textit{a priori} unseen in the training dataset.
    Our \acronym{} framework is able to predict the hyperplane structure of the solutions and also model the distribution shift of the planes with respect to the maximum allowable thrust $\alpha'$.
    }
    \label{Fig: AmorGS prediction hyperplane histogram}
\end{figure}
\FloatBarrier
We also find out that different hyperplane contains qualitatively different trajectories.
Figure \ref{Fig: trajectory} shows samples of trajectory solutions $\bm{x}^*(\alpha)$ with $\alpha = 1.0$ in three different hyperplanes within the CR3BP canonical reference frame.
The trajectory (shown in orange) starts from the end of an LT spiral (shown in cyan) and ends at a targeted arc (shown in green) in the stable invariant manifold (shown in black) of a Lyapunov orbit around $\mathcal{L}_1$ Lagrange point.
From the bottom left to the top right hyperplane, the sum of the initial and final coast times and shooting time increases, and the reason is that the trajectories contain more and more revolutions.

\textbf{Control Generation Model}

In this CR3BP example, as the time variables $\Delta T_{shooting}, \Delta T_{coast}^{initial}, \Delta T_{coast}^{final}$ are grouped into hyperplanes, the control variables $\bm{u}_1, \bm{u}_2, ..., \bm{u}_N$ as defined in Eq. \eqref{eq: cr3bp variable} are high-dimensional and much more complex, not showing explicit clustering patterns.
In this case, we only use the CVAE model to predict the time variable $\Delta T_{shooting}, \Delta T_{coast}^{initial},\Delta T_{coast}^{final}$ and final mass $m_f$.
Then to generate corresponding control solutions $\bm{u}_1, \bm{u}_2, ..., \bm{u}_N$, we propose an Auto-Encoder (AE) based Control Generation Model that maps from the time $\Delta T_{shooting}, \Delta T_{coast}^{initial}, \Delta T_{coast}^{final}$, final mass $m_f$, and the maximum allowable thrust $\alpha$ to the corresponding control solutions $\bm{u}_1, \bm{u}_2, ..., \bm{u}_N$.
\FloatBarrier
\begin{table}[!htbp]
\centering
\begin{subtable}{.45\linewidth}
\centering
\footnotesize
\begin{tabular}{p{2.2cm}cc}
\hline
 & \textbf{AmorGS} & \textbf{Uniform}\\
\hline
Count & $\bm{67}$ & 35\\
\hline
Solving time (s) \\
Mean & $\bm{59.18}$ & 114.02\\
Min & $\bm{2.43}$ & 13.53\\
25\% & $\bm{12.34}$ & 66.68\\
50\% & $\bm{27.01}$ & 100.33\\
\hline
\end{tabular}
\caption{Maximum allowable thrust $\alpha'=0.15$}
\label{tab:statistics1}
\end{subtable}%
\hspace{0.05\linewidth} 
\begin{subtable}{.45\linewidth}
\centering
\footnotesize
\begin{tabular}{p{2.2cm}cc}
\hline
 & \textbf{AmorGS} & \textbf{Uniform}\\
\hline
Count & $\bm{81}$ & 42\\
\hline
Solving time (s) \\
Mean & $\bm{34.60}$ & 78.74\\
Min & $\bm{1.14}$ & 4.26\\
25\% & $\bm{6.42}$ & 28.66\\
50\% & $\bm{13.32}$ & 59.74\\
\hline
\end{tabular}
\caption{Maximum allowable thrust $\alpha'=0.85$}
\label{tab:statistics2}
\end{subtable}
\caption{Statistical summary of locally optimal \textit{good} solutions ($m_f > 415$kg) and the respective solving time from 100 initializations for the minimum-fuel low-thrust transfer in the CR3BP. 
\acronym{} method warm-starts the global search from structured predictions and obtained more locally optimal \textit{good} solutions than the Uniform method, whose initializations are drawn from a uniform distribution.
Moreover, the solving time of these locally optimal solutions is significantly shorter for the \acronym{}. }
\label{tab:statistics}
\end{table}
\FloatBarrier
We display the framework of the CVAE in Figure \ref{fig:cvae architecture} and the Control Generation Model in Figure \ref{fig:control generation model architecture} for this minimum-fuel LT transfer problem.
The CVAE architecture is the same as we previously used for De Jong's 5th function.
In the Control Generation Model, all the time variables $\Delta T$, final mass $m_f$ and the parameter $\alpha$ will be first encoded to a latent space and then decode to a control sequence $\bm{u}_1, \bm{u}_2, ..., \bm{u}_N$, both with the Deep Neural Networks.
Since the output control contains temporal correlation, for the decoder we choose a Long Short-Term Memory (LSTM) \cite{hochreiter1997long} to generate a sequential output.

The LSTM is a type of Recurrent Neural Network (RNN), which has a specific architecture that is well-suited for dealing with sequential data.
An RNN usually has hidden layers that form directed cycles, which effectively enables it to store information about past inputs in its internal state, therefore allowing it to exploit temporal dependencies within the data sequence.
LSTM is a special kind of RNN and is designed to alleviate the vanishing gradient problem encountered in traditional RNNs. 
LSTMs have the ability to selectively remember or forget information through a series of gating mechanisms, which makes them particularly effective for handling longer sequences.

\textbf{Experiment results and analysis}


The architecture of both the CVAE and the Control Generation Model used for the CR3BP is shown in Table \ref{table: CVAE architecture cr3bp}. 
We build these two neural networks with PyTorch \cite{paszke2017automatic}.
For this particular experiment, we choose a GMM prior that consists of 20 components, each of which is a 4-dimensional Gaussian with a diagonal covariance matrix.
For the Control Generation Model, we adopt a 3-layer bi-directional LSTM.
During the training time, we train the CVAE and the Control Generation Model together, and in the testing time, we first sample the predictions of the time $\Delta T$ and final mass $m_f$ conditioned on the maximum allowable thrust $\alpha$ using the CVAE, and then generate the corresponding control sequence $\bm{u}_1, \bm{u}_2, ..., \bm{u}_N$ using Control Generation Model.
\FloatBarrier
\begin{figure}[!htbp]
    \centering
    \includegraphics[width=0.85\linewidth]{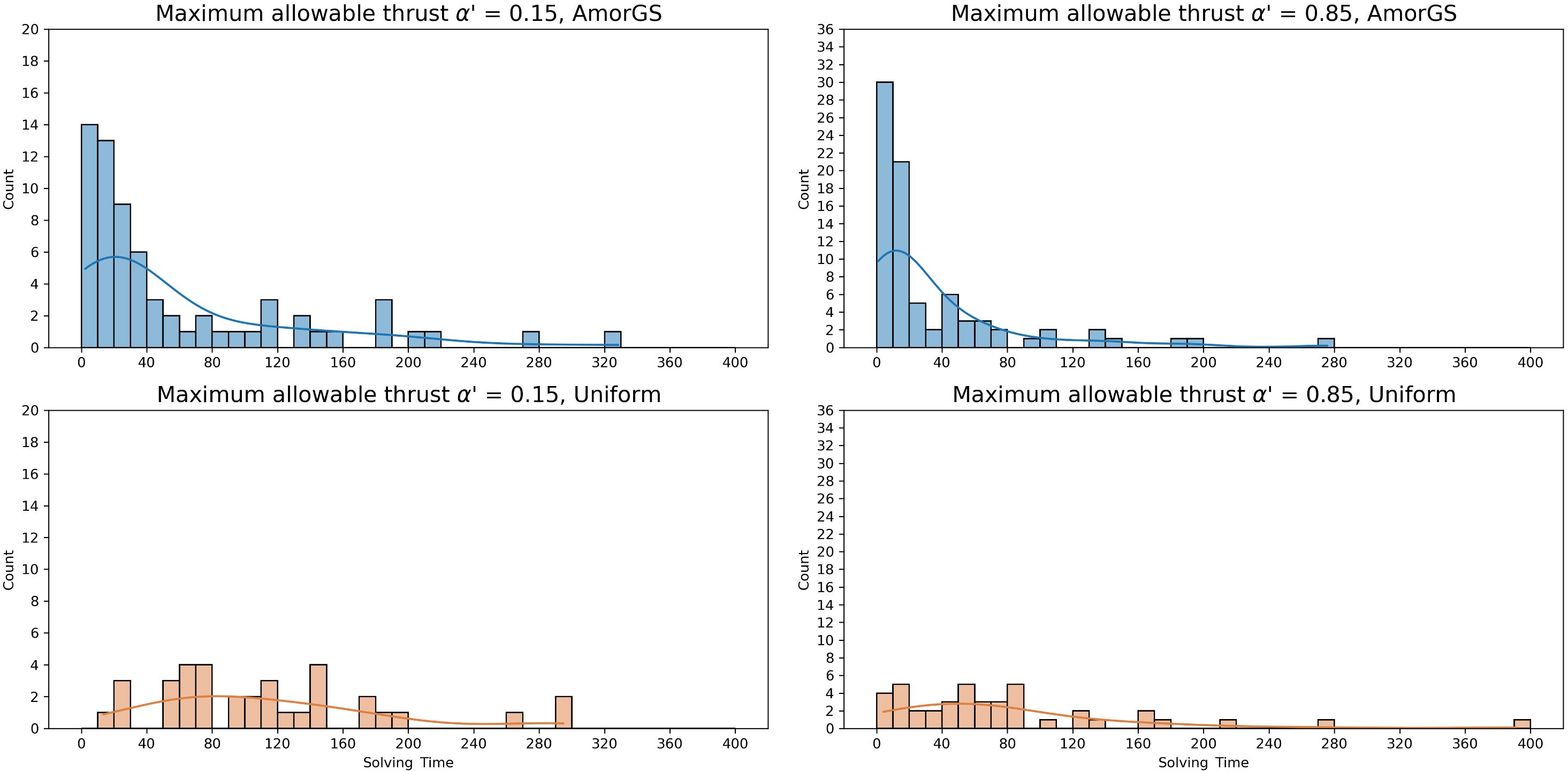}
    \caption{
    The histogram of solving time of locally optimal \textit{good} solutions ($m_f > 415$kg) to the minimum-fuel low-thrust transfer in the CR3BP within 100 initializations.
    \acronym{} warm-starts the global search with the predictions while the Uniform method uses uniformly sampled initializations.
    For $\alpha = 0.15$, \acronym{} obtains 27 locally optimal \textit{good} solutions within the first 20 seconds while the Uniform method only obtains 1.
    For $\alpha = 0.85$, \acronym{} obtains 30 locally optimal \textit{good} solutions within the first 10 seconds while the Uniform method only obtains 4.}
    \label{Fig: Snopt solving time histogram}
\end{figure}
\FloatBarrier
We test our method on minimum-fuel low-thrust transfer in the CR3BP with two maximum allowable thrust values $\alpha' = 0.15, 0.85$, which are unseen in the training data.
Figure \ref{Fig: AmorGS prediction hyperplane histogram} illustrates the hyperplane structure of time variables $\Delta T$ from our \acronym{} predictions alongside the histogram of the time-of-flight, compared to the groundtruth.
The blue plots denote our \acronym{} predictions and the orange plots are the groundtruth.
Our \acronym{} framework can accurately predict the hyperplane structure in the solutions and also has the ability to learn how the solution structure varies with respect to $\alpha'$.
As we move from a smaller $\alpha = 0.15$ to a larger $\alpha = 0.85$, our model's predictions shift toward areas with smaller time-of-flight, effectively mirroring the groundtruth.
This transition in solution density across each hyperplane is further illustrated in the histogram in Figure \ref{Fig: AmorGS prediction hyperplane histogram}.

Although the \acronym{} predictions themselves are already good estimations of the trajectory solutions, their feasibility and optimality are not guaranteed.
Therefore, we then use predictions from \acronym{} as initializations to warm-start the global search of the problem $\mathcal{P}_{\alpha'}$ with $\alpha' = 0.15$ and $0.85$.
This task poses a significant challenge: with uniformly sampled initial guesses, it requires a large number of initializations and extensive solving time for the solver $\pi$ to find a collection of qualitatively different and locally optimal \textit{good} solutions ($m_f \geq 415$kg).
In Table \ref{tab:statistics}, we present the number of locally optimal \textit{good} solutions $x^*(\alpha'), \alpha'=0.15, 0.85$ obtained from the solver $\pi$ with 100 initializations,  and the statistics of the respective solving time.
\acronym{} method warm-starts the solver with the structured prediction, while the Uniform method uses initializations drawn from a uniform distribution.
Within 100 initializations, \acronym{} yields 67 locally optimal \textit{good} solutions ($m_f > 415$kg), substantially outperforming the Uniform method which only produces 35.
More importantly, \acronym{} requires significantly less time to get these locally optimal \textit{good} solutions compared to the Uniform approach. 
Using \acronym{}, $50\%$ of the locally optimal solutions are obtained within the first 27 seconds for $\alpha' = 0.15$, and within the first 13 seconds for $\alpha' = 0.85$, which is approximately 4 times faster than the Uniform method.
\FloatBarrier
\begin{figure}[!htbp]
    \centering
    \includegraphics[width=0.9\linewidth]{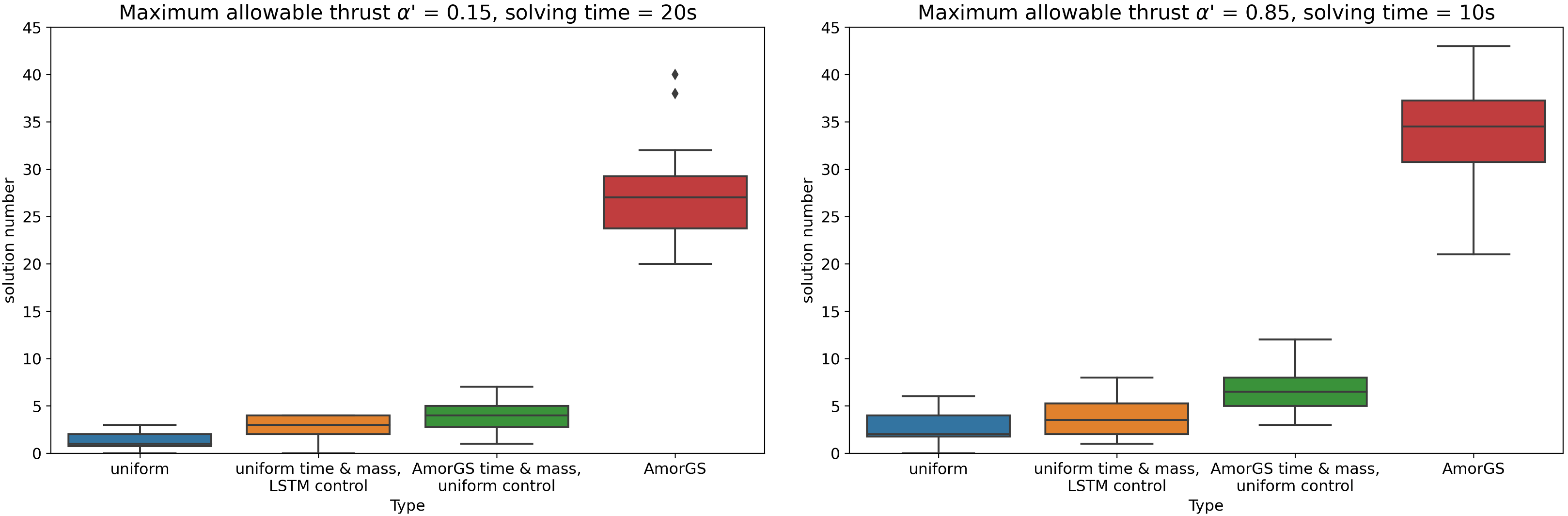}
    \caption{
    The box plot of locally optimal \textit{good} solutions ($m_f \geq 415$kg) obtained from 20 rounds of 100 initializations within only 20 seconds and 10 seconds of solving time for $\alpha'=0.15$ and $0.85$, respectively, to the minimum-fuel low-thrust transfer in the CR3BP.
    The box represents the data from the first quartile (Q1) to the third quartile (Q3) with the whiskers extending to 1.5 times the interquartile range and black dots as outliers. 
    From left to right, uniformly sampled variables are gradually replaced with predictions from different modules of \acronym{}, and the performance increases incrementally, but only by a small margin.
    The rightmost red box indicates that using the full \acronym{} framework, we can obtain significantly more solutions with a short solving time.}
    \label{Fig: Snopt 10s solution number box plot}
\end{figure}
\FloatBarrier
Figure \ref{Fig: Snopt solving time histogram} presents a histogram of solving time for locally optimal \textit{good} solutions across these 100 initializations.
We can see that \acronym{} method generates more  locally optimal \textit{good} solutions than the Uniform method and, does so in significantly less time.
Specifically, for $\alpha'=0.15$, \acronym{} can obtain 27 locally optimal \textit{good} solutions within the first 20 seconds, while the Uniform can only get 1.
Similarly for $\alpha'=0.85$, \acronym{} is able to get 30 locally optimal \textit{good} solutions within the first 10 seconds, while the Uniform method only gets 4.

Based on these observations, we set a cut-off time for the solver $\pi$ at 20 seconds for $\alpha'=0.15$ and 10 seconds for $\alpha'=0.85$.
Our goal is to improve the efficiency of the global search by spending less time per initialization but solving with more initializations overall.
We implement 20 rounds of global search, each with 100 initializations generated using different seeds.
The number of locally optimal \textit{good} solutions ($m_f \geq 415$kg) per round is represented in a box plot shown in Figure \ref{Fig: Snopt 10s solution number box plot}.
In the box plot, the box represents data from the first quartile (Q1)  to the third quartile (Q3), the whiskers extend to 1.5 times the interquartile range, and the black dots are outliers beyond the whiskers.
We conduct an ablation study to analyze the impact of different components used within the \acronym{} framework.
From left to right, Figure \ref{Fig: Snopt 10s solution number box plot} presents the types of initializations we tested: 1) all variables drawn uniformly, 2) time $\Delta T$ and final mass $m_f$ variables drawn uniformly, with control $\bm{u}$ provided by the Control Generation Model, 3) time $\Delta T$ and final mass $m_f$ variables provided by the CVAE in \acronym{}, with control $\bm{u}$ drawn uniformly, and 4) all variables provided by \acronym{} predictions.
Our analysis reveals that replacing uniformly drawn initializations with predictions from various \acronym{} modules slightly increases the number of locally optimal \textit{good} solutions.
However, a substantial performance increase becomes evident only when the complete \acronym{} framework is employed, as depicted by the red box.

Though many solutions can be obtained within a short solving time, we have to make sure that they are qualitatively different solutions, instead of solutions that just converge quickly.
Figure \ref{Fig: Amorgs accelerated solutions, work flow} depicts the entire result flow of our \acronym{}, for the accelerated global search on a minimum-fuel low-thrust transfer in the CR3BP with \textit{a priori} unseen maximum allowable thrust values $\alpha' = 0.15, 0.85$.
On the left of Figure \ref{Fig: Amorgs accelerated solutions, work flow}, we present the predictions of locally optimal \textit{good} solutions from \acronym{}.
These predictions already present a hyperplane structure for the time variables $\Delta T$, but with some noises.
Then the predictions are served as initializations to warm-start the global search, with a 20-second cut-off time for $\alpha' = 0.15$ and a 10-second cut-off time for $\alpha' = 0.85$ for the solver $\pi$.
As shown in the middle of Figure \ref{Fig: Amorgs accelerated solutions, work flow}, the accelerated solutions obtained by the solver exhibit a clearer hyperplane structure that closely mirror the groundtruth on the right.
This proves that even though the problem is solved in just a short period of time, we can still obtain qualitatively different solutions, benefiting from the speed-up performance provided by the \acronym{} framework.

\FloatBarrier
\begin{figure}[!htbp]
    \centering
    \includegraphics[width=.85\linewidth]{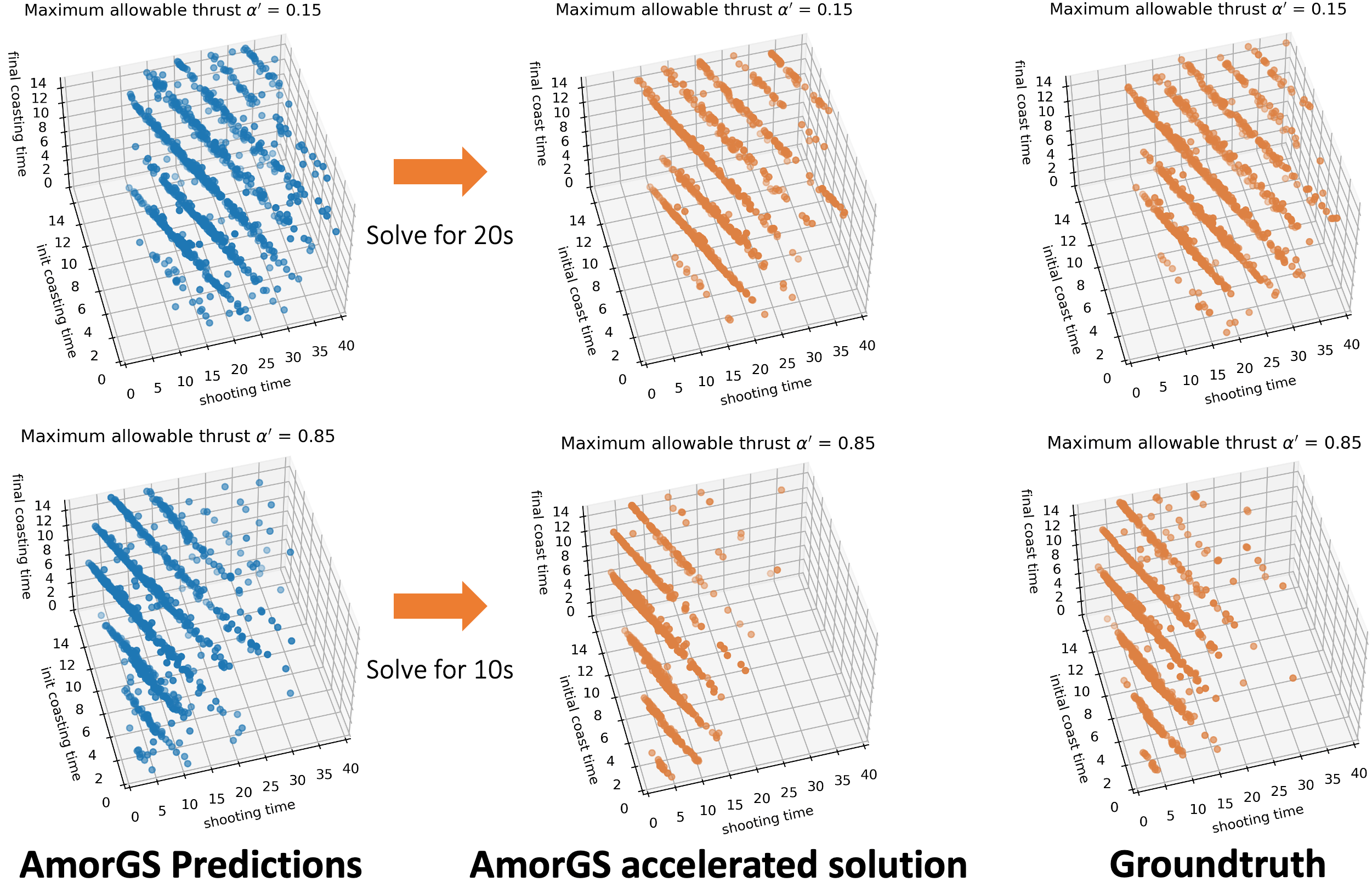}
    \caption{
    The result flow of \acronym{} on accelerating global search over solutions to a minimum-fuel low-thrust transfer in the CR3BP, with unseen maximum allowable thrust $\alpha'=0.15, 0.85$.
    The leftmost figure displays the initial \acronym{} predictions of the time variables, which already exhibit hyperplane structures but with some noises. 
    These predictions are then used as initializations to warm-start the global search.
    With only 20 seconds of solving time for $\alpha'=0.15$ and 10 seconds for $\alpha=0.85$, the solver $\pi$ is able to obtain qualitatively different solutions while enjoying the speed-up performance from \acronym{}, as shown in the middle plot. These accelerated solutions maintain a similar solution structure as the groundtruth shown on the right.
    }
    \label{Fig: Amorgs accelerated solutions, work flow}
\end{figure}
\FloatBarrier

\section{Conclusion}

In this paper, we introduce \acronym{}, a novel amortized global search framework for efficient preliminary trajectory design for spacecraft.
We discover that solutions to this highly non-convex trajectory optimization problem group into clusters that vary with respect to the problem parameter values, for instance, the maximum allowable thrust.
Using deep generative models, our proposed \acronym{} can predict the solution structure of a new problem instance with unseen parameter values, based on the solution data from similar problems.
The prediction ability and the acceleration performance are demonstrated on De Jong's 5th function and a minimum-fuel transfer in the CR3BP.

In the future, we will explore other generative models, e.g. the diffusion model \cite{ajay2022conditional, botteghi2023trajectory}, to build a unified framework for trajectory optimization problems.
The goal is to achieve successful multi-model predictions for both time variables exhibiting hyperplane structure and control variables with temporal correlations at the same time.
Furthermore, we will investigate how to refine the predictions of our framework using data gathered from the online global search process.
This becomes particularly crucial during a global search for a problem where the parameter values deviate significantly from the training data.
In such a scenario, the ability to gradually fine-tune our prediction model in \acronym{} based on a few shots of the groundtruth would be highly beneficial.

\section{Acknowledgement}

The simulations presented in this article were performed on computational resources managed and supported by Princeton Research Computing, a consortium of groups including the Princeton Institute for Computational Science and Engineering (PICSciE) and the Office of Information Technology's High-Performance Computing Center and Visualization Laboratory at Princeton University.

\bibliographystyle{AAS_publication}   
\bibliography{main.bbl}

\end{document}